\documentclass[journal]{IEEEtran}
\usepackage{graphicx}                                
\usepackage{amsmath}
\usepackage{multirow}
\usepackage{algorithm}
\usepackage{algorithmic}
\usepackage{cite}
\usepackage{amsthm}
\usepackage{amssymb}
\usepackage{array}
\usepackage{setspace}
\usepackage{booktabs}
\usepackage{float}
\usepackage{threeparttable}
\usepackage{caption,subcaption}
\usepackage{hyperref}
\usepackage{diagbox}
\usepackage{amsmath} 
\usepackage{multirow}
\usepackage[normalem]{ulem}
\useunder{\uline}{\ul}{}
\usepackage[table,xcdraw]{xcolor}

\hypersetup{hidelinks,
	colorlinks=true,
	allcolors=black,
	pdfstartview=Fit,
	breaklinks=true}

\begin{document}
%
\title{Dual Branch Neural Network for Sea Fog Detection in Geostationary Ocean Color Imager}
%
%
%
\author{
	Yuan Zhou,~\IEEEmembership{Senior Member,~IEEE,}
	Keran Chen,
	Xiaofeng Li,~\IEEEmembership{Fellow,~IEEE}

\thanks{This work was supported by the National Natural Science Foundation of China (U2006211 and 62171320) and National Key Research and Development Program of China (2020YFC1523204)} 
\thanks{(\emph{Corresponding author: Xiaofeng Li.})}

\thanks{Yuan Zhou and Keran Chen are with the School of Electrical and Information Engineering, Tianjin University, Tianjin 300072, China (e-mail: zhouyuan@tju.edu.cn; erichen@tju.edu.cn).}

\thanks{Xiaofeng Li is with Key Laboratory of Ocean Circulation and Waves, Institute of Oceanology, Chinese Academy of Sciences, and Center for Ocean Mega-Science, Qingdao, 266071 China. (email: xiaofeng.li@ieee.org).}
}

\markboth{IEEE Transactions on Geoscience and Remote Sensing,Vol.XX, No.XX, XXXX}%
{Shell \MakeLowercase{\textit{et al.}}: Bare Demo of IEEEtran.cls for IEEE Journals}

\maketitle
\begin{abstract}
	Sea fog significantly threatens the safety of maritime activities. This paper develops a sea fog dataset (SFDD) and a dual branch sea fog detection network (DB-SFNet).  We investigate all the observed sea fog events in the Yellow Sea and the Bohai Sea (118.1°E-128.1°E, 29.5°N-43.8°N) from 2010 to 2020, and collect the sea fog images for each event from the Geostationary Ocean Color Imager (GOCI) to comprise the dataset SFDD. The location of the sea fog in each image in SFDD is accurately marked. The proposed dataset is characterized by a long-time span, large number of samples, and accurate labeling, that can substantially improve the robustness of various sea fog detection models. Furthermore, this paper proposes a dual branch sea fog detection network to achieve accurate and holistic sea fog detection. The poporsed DB-SFNet is composed of a knowledge extraction module  and a dual branch optional encoding decoding module. The two modules jointly extracts discriminative features from both visual and statistical domain. Experiments show promising sea fog detection results with an F1-score of 0.77 and a critical success index of 0.63. Compared with existing advanced deep learning networks, DB-SFNet is superior in detection performance and stability, particularly in the mixed cloud and fog areas.
\end{abstract}


\IEEEpeerreviewmaketitle

\section{Introduction}

\IEEEPARstart{S}{ea} fog is a common and disastrous weather phenomenon when water vapor near the sea surface is condensed to form suspended water droplets \cite{leipper_fog_1994}. In foggy sea areas, horizontal visibility is less than 1 km, thereby highly threatening the safety of maritime activities such as shipping, aquaculture, among others \cite{zhu_demonstration_2017, pye_acidity_2020, riemer_aerosol_2019}. Therefore, the accurate detection of sea fog is a highly demanding and significant task. 

Yellow Sea and the Bohai Sea are two marginal seas between mainland China and the Korean peninsula in the western Pacific Ocean. Sea fog often occurs in these areas adjacent to the land on three sides. This area's primary type of sea fog is the advection fog formed by warm and humid air flowing on the underlying cold surface. In addition, there are many important trade routes in the Yellow Sea and the Bohai Sea. Frequent sea fog threatens navigation safety and trade stability. Therefore, accurate sea fog detection in this area is necessary. 

Data observed at meteorological stations, ships, and buoys have been traditionally used for sea fog detection. However, such observation data is sparse in spatial and temporal distribution \cite{kim_geostationary_2020}. Therefore, satellite remote sensing technology has facilitated sea fog observation over a large area in the past three decades \cite{mahdavi_probability-based_2021}. 

Existing sea fog detection is primarily based on physical models. By theoretically studying the lower emissivity of tiny droplets at mid-infrared and thermal infrared bands, Hunt \cite{hunt_radiative_1973} has concluded that differences in the radiative characteristics between fog types can be used as criteria in fog detection. Among these radiative characteristics, brightness temperature is the most distinctive for sea fog detection \cite{underwood_multiple-case_2004, gultepe_microphysical_2007, shin_new_2018}. However, solar radiation interfered with brightness temperature on the thermal infrared band and thus cannot be used for daytime sea fog detection.

In response to the daytime sea fog detection challenge, scientists analyzed the difference between spectra bands and developed a series of spectra-dependent detection algorithms. With the moderate-resolution imaging spectroradiometer (MODIS) data on polar-orbiting satellites Aqua and Terra, Deng et al. \cite{deng_detection_2014} set fixed thresholds to segment sea fog. Wu and Li \cite{wu_automatic_2014} proposed a dynamic threshold setting method to detect sea fog in the changing light conditions for MODIS data. Using the Cloud-Aerosol Lidar with Orthogonal Polarization (CALIOP) data based on vertical-resolved measurements, Wu et al. \cite{wu_method_2015} used the difference between the laser passing through the cloud and the fog to optimize the sea fog detection. 

Unlike polar-orbiting satellites, which provide only one or two images of the same geographic area per day, geostationary satellite sensors collect images at high temporal resolution. Therefore, there are also some researches on sea fog detection based on geostationary satellites. Based on Multi-functional Transport Satellite (MTSAT) data, Heo et al. \cite{heo_algorithm_2014} used the differences between dual-channel and texture to study the Korean Peninsula sea fog. Based on Geostationary Operational Environmental Satellite (GOES) data, Lee \cite{lee_stratus_1997} presented a method to detect fog continuously that used shortwave infrared data during the day.

The methods mentioned above are primarily linear algorithms based on thresholds; thus, such methods have limitations in complicated nonlinear scenarios. Recently, owing to its nonlinear fitting ability, deep learning technology has shown promising potential in oceanographic research \cite{li_deep-learning-based_2020, zheng_purely_2020, ren_development_2022}.

However, there are only a few studies on deep-learning-based sea fog detection. Zhu et al. \cite{zhu_sea_2019} used 16 samples to train a U-Net model \cite{zhou_unet_2020} for sea fog detection. Due to the scarcity of labeled samples, Huang et al. \cite{huang_correlation_2021} proposed a sea fog detection method that uses GAN \cite{choi_self-ensembling_2019} to generate enhanced sea fog images to compensate for the insufficient samples. 

The lack of labeled data limits the performance of deep learning sea fog detection studies. The training of the deep learning models requires large-scale labeled datasets. However, the existing research does not provide any large-scale data set to support the deep learning-based sea fog detection methods. 

Moreover, sea fog detection is more complicated than other segmentation tasks in the computer vision because of the difficulty in separating clouds and fogs. There are statistical differences between sea fog and cloud from the remote sensing image. Therefore, it is reasonable to detect sea fog from statistical differences. The aforementioned deep-learning-based methods directly applied existing deep learning models to sea fog detection tasks without considering the visual and statistical characteristics of the fog area. 

To solve the above issues, we first built a dataset and then designed a deep learning model that is more suitable for sea fog detection. The main contributions of this paper are summarized as follows:

\begin{enumerate}[]
	\item We propose a new sea fog detection dataset (SFDD) on remote sensing data. 
	We selected 1032 sea fog images recorded by the Geostationary Ocean Color Imager (GOCI) over the Yellow and Bohai through collecting, sorting, and careful annotation of many historical sea fog events seas from 2010 to 2020. The prerequisite for a deep learning model to be effective in real scenarios is that the distribution of the data set should be consistent with the distribution in the actual scenario. Therefore, the dataset is constructed considering the seasonal distribution and geographical distribution consistency between SFDD and historical statistics. Statistical studies show that the data distribution of SFDD exhibits adequate consistency with that of the actual scenes.
	
	\item We designed a dual branch sea fog detection model (DB-SFNet). 
	The two branches extract features from visual representation and statistical characteristics individually. In addition, a knowledge extraction module (KEM) is designed to  represent the pixel-level neighbor relation statistical characteristics to accurately detect sea fog events in complex scenes.
\end{enumerate}

The remainder of this paper is organized as follows. In Section II, we describe the proposed sea fog dataset. Section III introduces the preliminary related to our work. Section IV describes the method. Section V analyzes the detection results of DB-SFNet. Finally, this paper is concluded in Section VI.

\section{Construct Sea Fog Satellite Image Data Set}

A dataset with rich categories, sufficient data and accurate annotation is the cornerstone of deep learning-based sea fog detection algorithm. In the past, annotations for sea fog areas were station-level. That is, the sea fog was only sparsely recorded by individual ocean sites or buoys. Compared to these station-level labels, the labels in SFDD are pixel-level annotated based on satellite imagery and weather forecast reports. To the best of our knowledge, the proposed SFDD is the first sea fog detection dataset with an extended period, large sample number, and pixel-level labeling. 

\subsection{Satellite Data}

\begin{figure*}[!t]
	\centering
	\includegraphics[width=\linewidth]{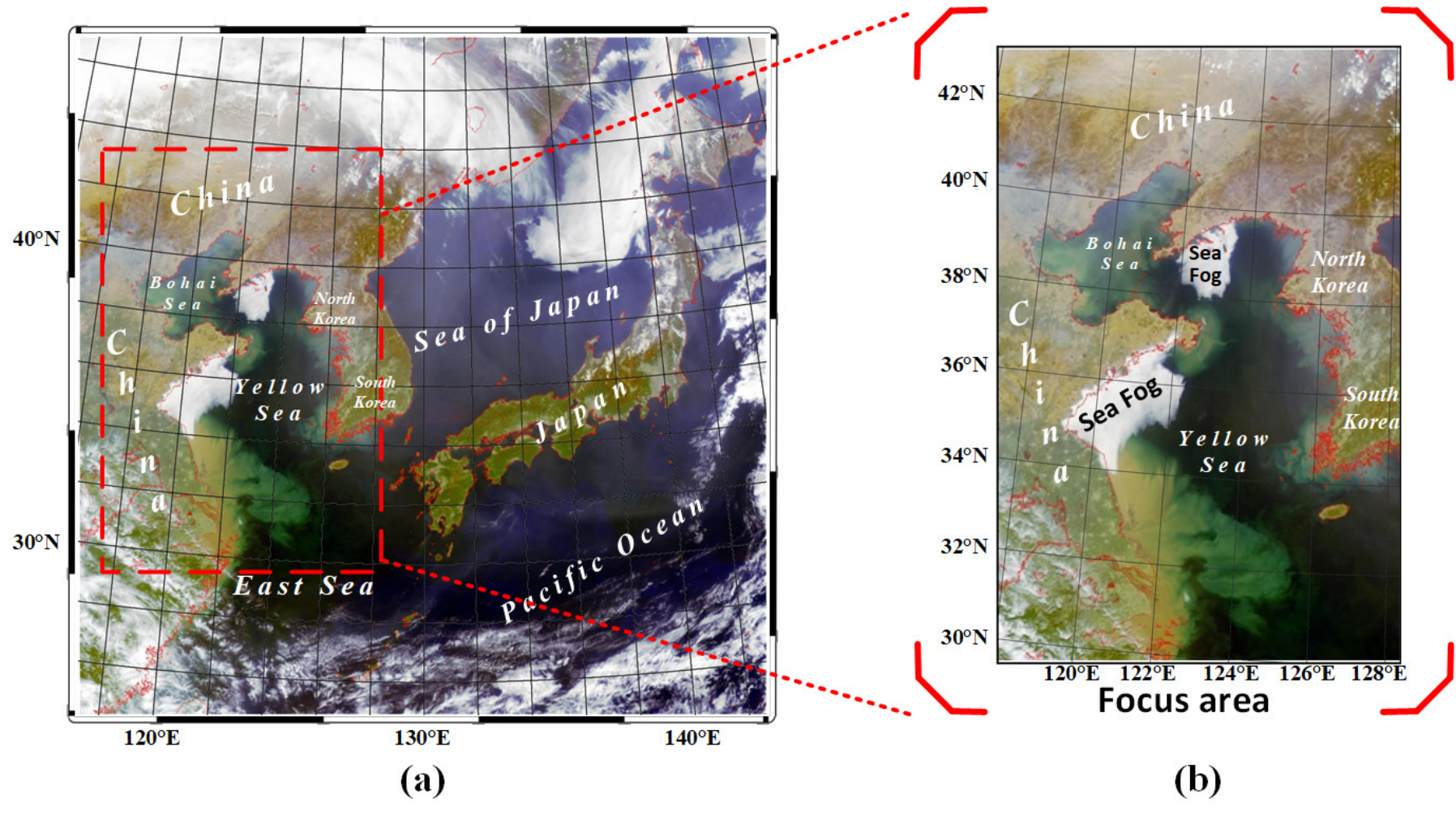}
	\caption{Study area. \textbf{(a) GOCI target area.} The spatial resolution of GOCI is approximately 500m and the range of target area is approximately 2,500 km×2,500 km, centered on the Korean Peninsula. \textbf{(b) Focused area.} The Yellow and Bohai seas, surrounded by China and the Korean Peninsula.}
	\label{Fig1}
\end{figure*}

The satellite data used to construct the dataset is the Level-2C multispectral data from GOCI. GOCI is one of the three payloads onboard the Communication, Ocean and Meteorological Satellite (COMS). It is the world's first geostationary orbit satellite image sensor used for the observation or monitoring of ocean color around the Korean Peninsula. The spatial resolution of GOCI is 500m and the imaging area is 2,500 km×2,500 km centered on 130°E, 36°N (shown in Fig.\ref{Fig1}(a)). GOCI provides satellite images at hourly intervals up to 8 times a day from 00:16 UTC to 07:16 UTC, allowing observations of short-term changes in the northeast Asia region. Six visible bands and two near-infrared bands are provided by GOCI. The region of interest area (118.1°E-128.1°E, 29.5°N-43.8°N) is shown in Fig.\ref{Fig1}(b).

\subsection{Construction of sea fog detection dataset}

The National Satellite Meteorological Centre (NSMC) of China has recorded fog events over the past ten years. During the ten years from 2011 to 2020, 133 sea fog incidents occurred in the Yellow Sea and the Bohai Sea. Based on each record, we downloaded GOCI Level-2C remote sensing images (after radiation correction, geometric correction and atmospheric correction) on the corresponding date. From 00:00 UTC to 08:00 UTC, GOCI takes one shot every hour and records eight bands of spectral information for each shot. We selected 490nm, 550nm, and 680nm wavebands as the red, green and blue channels to form color images. The region of interest area (118.1°E-128.1°E, 29.5°N-43.8°N) is cropped from the color image. 

The fog inversion product released by NSMC is used as ground truth. In this product, fog areas are manually labeled by meteorological experts, according to the fog product inversion manual of the NSMC. The labeling process is as follows.

\begin{enumerate}[]
	\item L1B level reflectance data is obtained from Fengyun Geostationary satellite; 
	
	\item Sea fog inversion is performed on reflectance data and terrain data using a threshold-based inversion algorithm; 
	
	\item Meteorologists correct the inversion results of Step 2 by using the visibility information recorded by the observation stations;
	
	\item Meteorologists make a second correction based on their experience.
\end{enumerate}

As a fog-monitoring product that has been applied in business for many years, its sea fog inversion results are highly reliable to be the ground truth labels. 

Three meteorological experts were hired to manually annotate the GOCI data using the fog inversion product of NSMC, according to the latitude and longitude. Each expert independently completed the annotation without interference from the others. The winner-take-all strategy was used to generate a more reliable label mask as the ground truth for error analysis.

\subsection{Statistical analysis of sea fog detection dataset}

\begin{figure}[t]
	\centering
	\includegraphics[width=\linewidth]{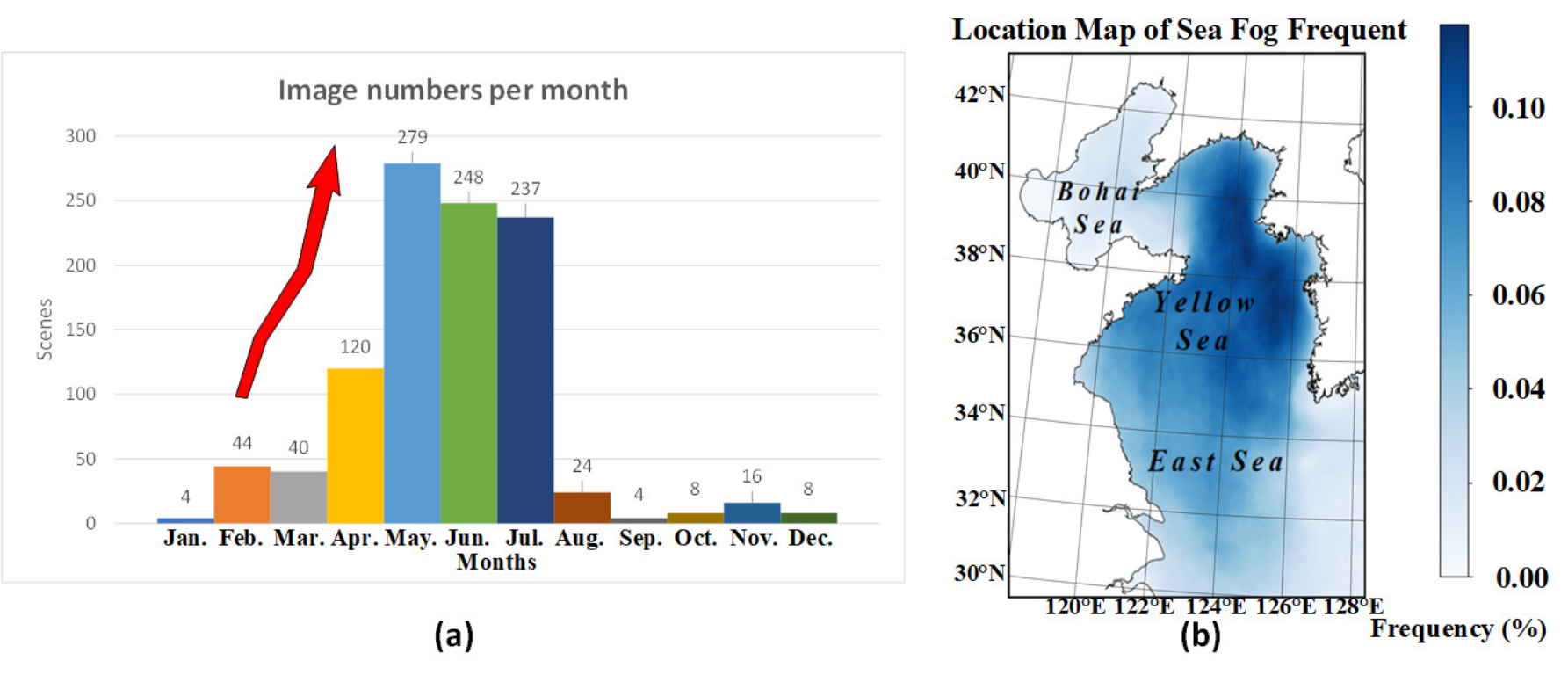}
	\caption{Statistical properties of SFDD. \textbf{(a) Monthly statistics of sea fog images.} Average monthly number of sea fog images collected by SFDD in the past ten years \textbf{(b) Location Map of Sea Fog Frequency.} Geographical location statistics with frequent occurrence of sea fog from 2011 to 2020. The darker the pixel color, the higher the frequency of sea fog in the corresponding geographic location in the past ten years.}
	\label{Fig2}
\end{figure}

We analyze to verify the consistency of data distribution for the proposed dataset. The statistical properties of the dataset are plotted in Fig. \ref{Fig2}. 

Fig. \ref{Fig2}(a) has been drawn by evaluating the frequency of sea fog occurrence each month. The high occurrence of sea fog in the Yellow Sea and the Bohai Sea starts in March and ends in July. The fog period is four months. Zhang et al. \cite{zhang_seasonal_2009} used observations from 15 standard weather stations between 1971 and 2000 to analyze the seasonal variations of sea fog in the Yellow Sea. The two main findings are: 1) The fog season onset is abrupt, with the number of fog days increasing rapidly from March to April. (As shown in Fig. \ref{Fig2}(a), the number of sea fog images has increased significantly since March.) 2) The frequency of sea fog incidence suddenly drops from its peak in July to almost zero in August. (The number of sea fog images in Fig. \ref{Fig2}(a) is still substantial in July, but it begins to drop suddenly in August.) This analysis demonstrates that the seasonal variation of the SFDD is consistent with historical statistics.

From spatial distribution, Fig. \ref{Fig2}(b) illustrates the distribution of sea fog in the Yellow Sea and the Bohai Sea during ten years. Wu et al. \cite{wu_analyses_2015} analyzed sea fog distribution and seasonal characteristics in the Yellow Sea and the Bohai Sea from 1989 to 2008. They concluded that 1) in the Bohai Sea, only the coasts of the Liaodong Peninsula and the Shandong Peninsula have sea fog, and  2) sea fog appears most frequently in the central part of the Yellow Sea and the West Korea Bay. These conclusions on the spatial distribution of sea fog are consistent with the result shown in Fig. \ref{Fig2}(b). 
 
The statistical information of SFDD is consistent with the conclusion of \cite{zhang_seasonal_2009} and \cite{wu_analyses_2015}, indicating that the data distribution of SFDD is closer to the natural data distribution. A model trained on the SFDD dataset is expected to demonstrate an improving generalization ability.

\section{Preliminary: Fully Convolutional Network}

\begin{figure}[h]
	\centering
	\includegraphics[width=\linewidth]{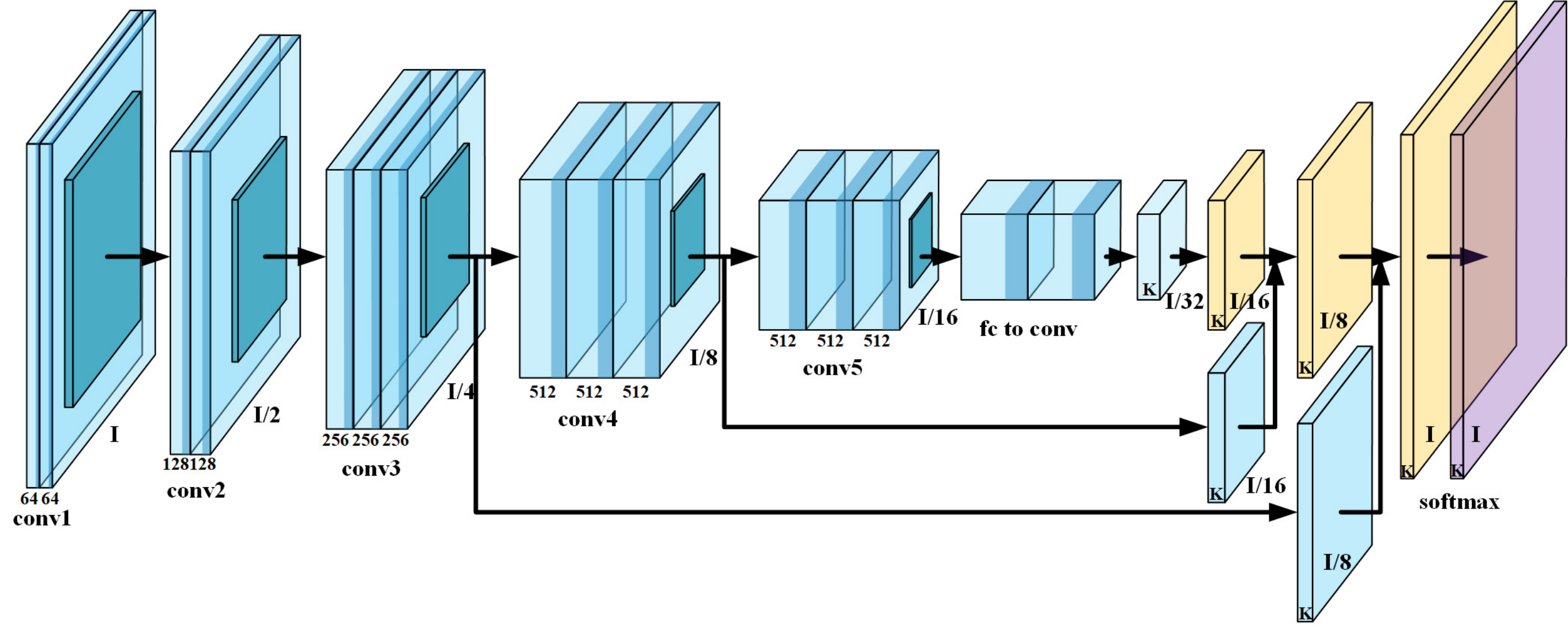}
	\caption{The structure of Fully Convolution Network.}
	\label{Fig3}
\end{figure}

Sea fog detection is usually considered a segmentation task in the computer vision community. The fully convolutional network (FCN) proposed by Long et al. \cite{long_fully_2015} is the classical deep learning method for segmentation tasks. It is capable of achieving pixel-wise classifications in the sea fog detection task. The structure of FCN is shown in Fig. \ref{Fig3}. 

As shown in Fig. \ref{Fig3}, an FCN network consists of one encoder and one decoder. A series of stacked convolutional layers constitute the encoder. The first layer takes the image ($x_i\in R^{I\times I\times3}$) as the input. The inputs of the subsequent layer are 3D feature maps sized $\frac{I}{n} × \frac{I}{n} × d$, where $n$ is the scaling factor of the feature map, and $d$ denotes the channel dimension. After the original encoder reduces the feature map to $1/32$ of the original image, the decoder restores the feature map to the size of the original image through convolution and upsampling layers. By using the paradigm of encoding and decoding, FCN can realize the foreground detection of natural images. During detection, the convolutional layer extracts discriminator features from the visual representation. 
 
Statistical features are also important when detecting sea fog from remote sensing images mixed with clouds and fog. Therefore, by retaining the encoding and decoding paradigm of FCN, we design a dual-branch network for sea fog detection tasks that can encode features from both visual and statistical spaces. In addition, to avoid the problem of complex decoding caused by the large number of features encoded from the two spaces, we designed a feature selection module for the decoder.

\section{Proposed Method}

\subsection{Network Architecture}

\begin{figure*}[tbp]
	\centering
	\includegraphics[width=0.95\linewidth,height=0.72\textheight]{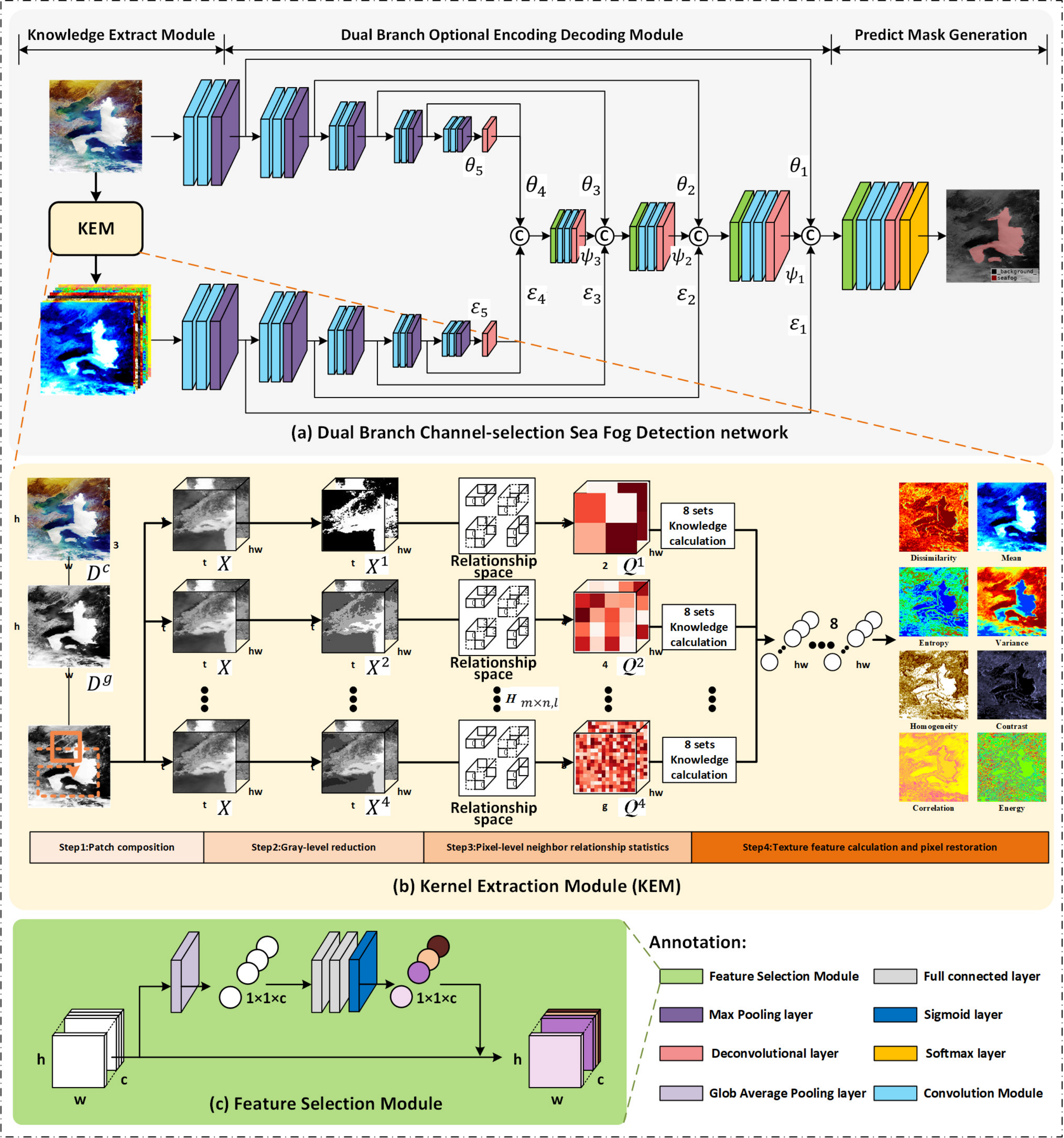}
	\caption{The architecture of the proposed DB-SFNet. DB-SFNet is divided into the knowledge extract module, dual branch optional encoding-decoding module, and predict mask generation. (a) Overall architecture. (b) Structure of the knowledge extraction module. (c) Structure of the feature selection module. }
	\label{Fig4}
\end{figure*}

A dual-branch sea fog detection network (DB-SFNet) is proposed here. Unlike the state-of-the-art semantic segmentation methods, DB-SFNet extracts information from visual representation domain and comprehensively considers the information of the statistical domain. A flowchart of the overall network is shown in Fig. \ref{Fig4}. DB-SFNet contains three main components: a knowledge extraction module (KEM), a dual-branch optional autoencoder (DOAE), and a prediction mask generation unit. The input of the proposed model is the true color band composition of GOCI, and the output is the detection result. KEM realizes the mapping of the input from the RGB domain to the statistical domain. The DOAE of the encoder-decoder structure comprises a dual-branch encoder and a decoder with feature selection. The prediction mask generation unit is responsible for pixel-level classification. The pixels classified as positive samples are detected as sea fog in the prediction map.

\subsection{Knowledge Extraction Module}

To distinguish between cloud and fog in the statistical space, we designed a KEM module to transform from visual space to statistical space. Using prior knowledge, KEM can extract statistical information in the visual space. The input and output of KEM are the images in visual space and statistical space, respectively. The structure of KEM is shown in Fig. \ref{Fig4}(b). The four main steps are explained below.

\textbf{Step 1: Patch composition.} As an input, the KEM receives an image $D\in R^{H\times W\times C}$, where $\left(H,W\right)$ is the resolution of $x$ , and $C$ is the number of channels. Let $d_{h,w,c}$ denote the pixel value of each color channel in the $h$-th row, $w$-th column, and $c$-th color channel of $D$.We first grayscale the image $D$ using the formula.

\begin{equation}
	d_{h,w}^g=\frac{1}{C}\sum_{c=0}^{C-1}d_{h,w,c}
\end{equation}
where $d_{h,w}^g$ is the pixel value of the grayscale image $D^g$ in the $h$-th row and $w$-th column, After that, we crop rectangular areas with length $t$ from $D^g$, taking each pixel of $D^g$ as the center points of those rectangular areas. Each rectangular area is regarded as a patch, and the subsequent steps will be performed on each patch.

\textbf{Step 2: Gray-level reduction.} The rectangular areas in grayscale image $D^g$ are stitched in the channel dimension to obtain a tensor $X\in R^{t\times t\times H W}$. Calculating the statistical features on a large gray level tensor is inefficient, therefore, the gray level of $X$ should be reduced by upsampling. As shown in Fig. \ref{Fig4}(b), $X^\alpha\in R^{t\times t\times H W} (\alpha =1,2,3,…,N)$ is the tensor obtained after the gray-level reduction, whereas $\alpha$ represents the value of the gray level. $x_{i,j,k}^\alpha$ and $x_{i,j,k}$ denote the value in the $i$-th row, $j$-th column and the $k$-th channel of the tensor $X^\alpha$ and $X$, respectively. We upsample the gray level of $X$ according to the formula
\begin{equation}
	x_{i,j,k}^\alpha=\left\lceil\left(x_{i,j,k}+1\right)\times2^{\alpha-8}\right\rceil-1
\end{equation}
where $\left\lceil\cdot\right\rceil$ denotes the round-up operation for each element in the tensor.

\textbf{Step 3: Pixel-level neighbor relationship statistics.} Statistical features are essentially the statistics of position and value relationships of every two pixels in $X^\alpha$.  From the position relationship analysis, there are four adjacent relationships between two adjacent pixels: the up and down, the left and right, and the two diagonal relationships. In Fig. \ref{Fig4}(b), four tensors represent these four relationships. From the value analysis, the value range of pixels is $[0, \alpha -1]$. Let $l\in\left[0,3\right]$ represent the position relationship between two adjacent pixels, $m,n\in\left[0,\alpha-1\right]$ represent the values of adjancent pixels. A relationship space is denoted as $H\in R^{\alpha\times\alpha,4}$, where the elements in $H$ are triads represented as $(m,n,l)$. The number of triads corresponding to $X^\alpha$ is $4\alpha^2$. 

Let $Q^\alpha$ denote the relationship matrix, whereas the coordinates are the values of the adjacent pixels in $H$ (i.e., $m$ and $n$), and $q_{m,n,k}^\alpha$ denotes the value in the $m$-th row, the $n$-th column, and the $k$-th channel of $Q^\alpha$. According to the principle of gray level co-occurrence matrix \cite{welch_cloud_1988}, $q_{m,n,k}^\alpha$ can be calculated using 
\begin{equation}
	\begin{small}
				q_{m,n,k}^\alpha\!=\!\frac{1}{4}\sum_{l=0}^{3}\sum_{i=1}^{t-1}\sum_{j=1}^{t-1}\\{(x_{i,j,k}^\alpha\!\land\!m)  \land(n \!\land\! x_{i+bias\left(l;0\right),j+bias\left(l;1\right),k}^\alpha)}
	\end{small}
\end{equation}
$bias\left(l;.\right)$ denotes the pixel shift according to the position relationship $l$. For $l=$0, 1, 2 and 3, $(bias\left(l;0\right)$,$bias\left(l;1\right))$ is set to $(0, 1)$, $(1,0)$, $(1, 1)$ and $(-1, 1)$, respectively. 

\textbf{Step 4: Statistical feature calculation and pixel restoration.} The confusion matrix $Q$ obtained in the previous step is the statistical information of the relationship between the pixel pairs at different gray levels. By selecting eight sets of knowledge calculation, we map the statistical information of the input to statistical features $F^S\in R^{HW}$ with\ elements\ $f_k^S$, where $S$ separates different statistical characteristics involving {mean, variance, homogeneity, contrast, dissimilarity, entropy, energy, correlation}. Values of these statistical characteristics are calculated according to standard statistical formulas \cite{welch_cloud_1988}.

To measure the average level of pixel values in the $t$-neighborhood of the center pixel, we calculate the mean feature $F^{mean}$ by
\begin{equation}
	f_k^{mean}=\frac{1}{N}\sum_{\alpha=1}^{N}\sum_{m,n=0}^{N^2-1}{i\times q_{m,n,k}^\alpha}
\end{equation}
where $Z$ represents the number of gray-level upsampling in step two, $q_{m,n,k}^\alpha$ denotes the statistical information in the $m$-th row and $n$-th at the patch centered on the $k$-th pixel at the $\alpha$-th grey level.
 
To measure how far the pixel value is spread out from their average value, we calculate the variance feature $F^{variance}$ by
\begin{equation}
	f_k^{variance}=\frac{1}{N}\sum_{\alpha=1}^{N}\sum_{m,n=0}^{N-1}{q_{m,n,k}^\alpha\times\left(i-f_k^{mean}\right)^2}
\end{equation}
The larger the variance of a patch, more discrete is the pixel distribution in this patch. 

The homogeneity feature $F^{homogeneity}$ is calculated to measure the extent of image statistical changes locally.
\begin{equation}
	f_k^{homogeneity}=\frac{1}{N}\sum_{\alpha=1}^{N}\sum_{m,n=0}^{N-1}\frac{q_{m,n,k}^\alpha}{1+\left(m-n\right)^2}
\end{equation}
Compared with clouds, the change in reflectance of sea fog areas is small, and the value of $F^{homogeneity}$ is higher.
 
The contrast feature $F^{contrast}$ is calculated to measure the clarity of the patch and the depth of the statistical groove. The formula is as follows:
\begin{equation}
	f_k^{contrast}=\frac{1}{N}\sum_{\alpha=1}^{N}\sum_{m,n=0}^{N-1}{q_{m,n,k}^\alpha\times\left(m-n\right)^2}
\end{equation}
In the remote sensing image, the statistical groove of the cloud is deeper, and the statistical groove of the fog is shallow. Therefore, $F^{contrast}$ of the point corresponding to the cloud is large, and the point corresponding to the fog is small.
 
Entropy feature $F^{entropy\ }$ and dissimilarity feature $F^{dissimilarity}$ are calculated so as to measure the degree of non-uniformity of patch. The formulas are as follows:
\begin{equation}
	f_k^{dissimilarity}=\frac{1}{N}\sum_{\alpha=1}^{N}\sum_{m,n=0}^{N-1}{q_{m,n,k}^\alpha\times\left|m-n\right|}
\end{equation}
\begin{equation}
	f_k^{entropy}=\frac{1}{N}\sum_{\alpha=1}^{N}\sum_{m,n=0}^{N-1}{q_{m,n,k}^\alpha\times l n\left(q_{m,n,k}\right)}
\end{equation}
Unlike clouds, the pixel distribution in the sea fog area in the multispectral image is more uniform, therefore the $F^{entropy\ }$ and $F^{dissimilarity}$ of the corresponding position are smaller.
 
We use energy feature $F^{energy}$ to measure the thickness of patch in the co-occurrence matrix.
\begin{equation}
	f_k^{energy}=\frac{1}{N}\sum_{\alpha=1}^{N}\sum_{m,n=0}^{N-1}\left(q_{m,n,k}^\alpha\right)^2
\end{equation}
A smaller $F^{energy}$ indicates lesser thickness of the sea fog.
 
We calculate the correlation feature $F^{correlation}$ to measure the local correlation in the image.
\begin{equation}
	f_k^{correlation}=\frac{1}{N}\sum_{\alpha=1}^{N}\sum_{m,n=0}^{N-1}{q_{m,n,k}^\alpha\times\frac{\left(m-f_k^{mean}\right)\left(n-f_k^{mean}\right)}{f_k^{variance}}}
\end{equation}
When the element values are uniformly equal, the $F^{correlation}$ is large; on the contrary, if the element values differ significantly, $F^{correlation}$ is small. In general, $F^{correlation}$ of the cloud is smaller than $F^{correlation}$ of the sea fog.

Finally, eight statistical features on each patch are calculated: mean, variance, homogeneity, contrast, dissimilarity, entropy, energy, and correlation. Then, each patch's eight statistical feature values are spliced according to the coordinates of the pixel to obtain the statistical space representation.

\subsection{Dual Branch Optional Encoding Decoding Network}

We design DOEN to extract features from $D$ and $F$. DOEN consists of two parts: an encoder and a decoder. The essence of the encoder is to extract the discriminative features of two spaces. The encoder is a dual-branch CNN. One branch encodes the features of the visual space, and the other branch encodes the statistical space features. The two branches have the same structure, but the parameters are not shared. Through the operation of convolution + maxpooling, coding features of different scales are extracted from $D$ and $F$. The structure of visual encoder branch is shown in Table\ref{table1}. These multiscale features are all involved in the decoding process.

\begin{table}[ht]
	\centering
	\caption{The structure of visual encoder branch}
	\label{table1}
	\begin{tabular}{lll}
		\hline
		\textbf{Layer} & \textbf{Output Shape} & \textbf{Params} \\ \hline
		InputLayer     & 3×256×256             & 0               \\
		Conv2D         & 64×256×256            & 1792            \\
		Conv2D         & 64×256×256            & 36928           \\
		MaxPooling2D   & 64×128×128            & 0               \\
		Conv2D         & 128×128×128           & 73856           \\
		Conv2D         & 128×128×128           & 147584          \\
		MaxPooling2D   & 128×64×64             & 0               \\
		Conv2D         & 256×64×64             & 295168          \\
		Conv2D         & 256×64×64             & 590080          \\
		MaxPooling2D   & 256×32×32             & 0               \\
		Conv2D         & 512×32×32             &                 \\
		Conv2D         & 512×32×32             &                 \\
		MaxPooling2D   & 512×16×16             & 0               \\
		Conv2D         & 512×16×16             & 2359808         \\
		Conv2D         & 512×16×16             & 2359808         \\
		MaxPooling2D   & 512×8×8               & 0               \\ \hline
		\multicolumn{3}{r}{\textbf{Total:5865024}}               \\ \hline
	\end{tabular}
\end{table}

The function of the decoder is to fuse the encoded features of different scales and restore them to the scale of the original image. We first fuse the smallest-scale features $\theta_5$ and $\varepsilon_5$. After that, the feature $\theta_{\omega-1}$ from the visual space and the feature $\varepsilon_{\omega-1}$ from the statistical space with the scale $\omega-1$ are fused by feature selection module, with the fused features $\psi_{\omega-1}$ from the previous scale $\omega-1$. As shown in Fig. \ref{Fig4}(c), the feature selection module is computed as 
\begin{small}
	\begin{equation}
		\psi_\omega=\mathcal{F}(GAP([\theta_{\omega-1},\psi_{\omega-1},\varepsilon_{\omega-1}]))\odot[\theta_{\omega-1},\psi_{\omega-1},\varepsilon_{\omega-1}]
	\end{equation}
\end{small}
where $GAP(\cdot)$ denotes the global average pooling, $F(\cdot)$ denotes a fully connected network, $\odot$ denotes the Hadamard product, and $[\cdot]$ is the concatenate operation in the matrix. A convolution and deconvolution layer follow each feature selection module to restore feature scales.

\subsection{Predict Mask Generation}

Through five sets of deconvolution feature selection modules and deconvolution operations, the obtained tensor $\Phi\epsilon R^{H,W,K}$ is with the same length and width as the original image, and the number of channels is the same as the number of prediction categories. In the sea fog detection task, the number of predicted categories $K=2$. The generation of the prediction mask $\mathrm{\Gamma}$ can be written as
\begin{equation}
	\Gamma\left( {i,j} \right) = \left\{ \begin{matrix}
		{0,~~\frac{e^{\Phi(i,j,1)}}{\sum_{k = 0}^{K}e^{\Phi(i,j,k)}} < thresh} \\
		{1,~~\frac{e^{\Phi(i,j,1)}}{\sum_{k = 0}^{K}e^{\Phi(i,j,k)}} \geq thresh} \\
	\end{matrix} \right.
\end{equation}
where thresh denotes the threshold that distinguishes sea fog from the background, the value of the threshold is discussed in Section V.C.

\section{Experimental Results}

This section comprehensively evaluates the proposed DB-SFNet on GOCI satellite thumbnails. Specifically, we first discuss data preparation and experimental settings. Then, we analyze the sea fog detection performance by DB-SFNet and other state-of-the-art methods. After that, we design ablation experiments showing that each module plays an influential role in detecting. Finally, we present case studies to show the real-world applications.

\subsection{Data Preparation and Experimental Settings}

\subsubsection{Data Set Partitioning} The model is trained on the observed true- color images and tested on the newly acquired true- color images from real scenarios. The newly acquired data is entirely different from the historical data. However, randomly dividing the dataset will lead to the training, and test sets will contain similar true- color images. The reason is that each daytime sea fog in the data set consists of eight consecutive true- color images required every hour. Therefore, randomly dividing the training and test sets will cause impractically high test performance.

To ensure the practicality of the test performance, we restrict the foreground distribution of the training set, verification set, and test set during datasets partitioning. The 1,040 sea fog images included in SFDD describe 133 sea fog events. Therefore, we randomly select 78 sea fog processes (616 images) as training sets, 27 processes (208 images) as validation sets, and 27 processes (208 images) as test sets.
 
\subsubsection{Data Augmentation} Three data augmentation methods (as follows) are chosen to compensate for the limited number of training sets. 

\begin{itemize}[]
	\item \textbf{Noise Injection:} Affected by the periodic drift of sensors and electromagnetic interference between components, remote sensing images are prone to noise. The typical noise in SFDD is the light spot. Therefore, 20\% of the images are contaminated with speckle-noise during the data augmentation;
	
	\item \textbf{Rotation:} The shape of the sea fog does not correlate with its position in the image. Therefore, rotation can increase the number of samples in the training set. Each image used to train exhibits a 50\% probability of -20° to 20° clockwise rotation;
	
	\item \textbf{Random Erasing:} Due to the occlusion of clouds, the structure of some parts of the observed sea fog is incomplete. To simulate a partial occlusion of the foreground, we randomly erase the training images. As a result, the foreground of each image has a 20\% probability of being blocked by the mask. 
\end{itemize}

\subsubsection{Evaluation Metrics} To comprehensive measure the sensitivity and specificity of the model, we utilize seven widely used quantitative metrics, that is, intersection-over-union (IoU), mean intersection-over-union (mIoU), accuracy, precision, recall, F1-score, critical success index (CSI), and kappa coefficient (Kappa).

\begin{table}[ht]
	\centering
	\caption{Confusion Matrix and Evaluation Metrics}
	\label{table2}
	\resizebox{0.45\textwidth}{21mm}{
	\begin{tabular}{cc
			>{\columncolor[HTML]{E2EFD9}}c ccc}
		& & \multicolumn{2}{c}{\cellcolor[HTML]{CCFFFF}Predicted Condition} & & \\[5pt]
		& \cellcolor[HTML]{EDEDED}Pixels   & \cellcolor[HTML]{CCFFFF}Positive($PP$) & \cellcolor[HTML]{CCFFFF}Negtive($PN$) &  & \\[5pt]
		\cellcolor[HTML]{FFF2CC}                                  & \cellcolor[HTML]{FFF2CC}\rotatebox[origin=c]{90}{Positive($P$)} & \begin{tabular}[c]{@{}c@{}}\textbf{True-positive}\\ $TP$\\ hit\end{tabular}                          & \cellcolor[HTML]{FBE4D5}\begin{tabular}[c]{@{}c@{}}\textbf{False-negative}\\ $FN$\\miss underestimation\end{tabular} & \cellcolor[HTML]{E2EFD9}\begin{tabular}[c]{@{}c@{}}Recall\\ $R=\frac{TP}{TP+FN}$\end{tabular}   & \\[30pt]
		\multirow{-5}{*}{\cellcolor[HTML]{FFF2CC}\rotatebox[origin=c]{90}{Actual Condition}} & \cellcolor[HTML]{FFF2CC}\rotatebox[origin=c]{90}{Negative($N$)}  & \cellcolor[HTML]{FBE4D5}\begin{tabular}[c]{@{}c@{}}\textbf{False-positive}\\ $FP$\\ false alarm\end{tabular}  & \cellcolor[HTML]{E2EFD9}\begin{tabular}[c]{@{}c@{}}\textbf{True-negative}\\ $TN$\\ correct rejection\end{tabular} &                                                                                 & \cellcolor[HTML]{E2EFD9}\begin{tabular}[c]{@{}c@{}}background IoU\\ $back\ IoU=\frac{TN}{TN+FP+FN}$\end{tabular} \\[30pt]
		&                                                                              & 
		\begin{tabular}[c]{@{}c@{}}Precision\\[7.5pt] $P=\frac{TP}{TP+FP}$\\[7.5pt] \end{tabular}                      &                                                                              & \cellcolor[HTML]{DEEAF6}\begin{tabular}[c]{@{}c@{}}Accuracy\\[7.5pt] $Acc=\frac{TP+TN}{TP+FN+TN+FP}$\\[7.5pt]\end{tabular} & \cellcolor[HTML]{DEEAF6}\begin{tabular}[c]{@{}c@{}}Kappa\\[7.5pt] $Kap=\frac{Acc\times n^2-P\times P P+N\times P N}{n^2-P\times P P+N\times P N}$ \\[7.5pt]\end{tabular}          \\
		&                                                                              & \begin{tabular}[c]{@{}c@{}}CSI\\[7.5pt] $CSI=\frac{TP}{TP+FP+FN}$\\[7.5pt]\end{tabular}                            &                                                                              & \cellcolor[HTML]{DEEAF6}\begin{tabular}[c]{@{}c@{}}F1Score\\[7.5pt] $f1=\frac{2\times p r e c i s i o n\times r e c a l l}{precision+recall}$\\[7.5pt]\end{tabular}  & \cellcolor[HTML]{DEEAF6}\begin{tabular}[c]{@{}c@{}}mean IoU\\[7.5pt] $mIoU=\frac{CSI+background\ IoU}{2}$\\[7.5pt]\end{tabular}      
	\end{tabular}}
\end{table}

Table \ref{table2} shows the confusion matrix and evaluation metrics of sea fog forecast results. The confounding matrix composed of $TP$, $FN$, $FP$ and $TN$ was obtained based on the statistical prediction results in pixel unit. The statistical content is:

\begin{itemize}[]
	\item $TP$: sea fog is existent and detected;

	\item $FN$: sea fog is existent but not detected;
	
	\item $FP$: sea fog is not existent but detected;
	
	\item $TN$: sea fog is not existent and not detected.
\end{itemize}

The green metrics in the table only evaluate the model's performance in detecting sea fog or background, that is, recall, precision, sea fog IoU, and background IoU. The blue metrics are the comprehensive evaluation of the model detection ability, accuracy, Kappa, F1-score, and mIoU.

\subsubsection{Experimental Setting} The DB-SFNet and all comparison segmentation methods were trained under the Tensorflow2.5.0 framework and optimized by the Adam algorithm\cite{kingma_adam_2017}. The operating system is Ubuntu 18.04, equipped with NVIDIA GTX 1080 Ti GPU. The proposed DB-SFNet is trained in an end-to-end manner. The learning rate started with 0.001, whose decay policy is cosine annealing \cite{loshchilov_sgdr_2017}. The minibatch size and total iterations are 12 and 8.66 × 103, respectively. In addition, the warmup \cite{xiong_layer_2020} strategy is used in the initial stage of training. Finally, the comparison methods are trained with the same parameter settings as those of the DB-SFNet and fine-tuned with their corresponding pre-trained CNN weights.

\subsection{Comparison with Advanced Segmentation Methods}

\subsubsection{Comparison Methods} Sea fog detection is a particular semantic segmentation task in computer vision. We compared our proposed DB-SFNet with three types of advanced segmentation models. The first category is convolution-based segmentation algorithms, including SegNet \cite{badrinarayanan_segnet_2017}, RefineNet \cite{lin_refinenet_2017}, PAN \cite{li_pyramid_nodate}, BiSeNet \cite{yu_bisenet_2018}, DenseASPP \cite{yang_denseaspp_2018}, and DeepLabV3+ \cite{chen_encoder-decoder_2018}. The second category contains three transformer-based segmentation algorithms, that is, Swin-Transformer \cite{liu2021swin}, Segmenter \cite{strudel2021segmenter} and Segformer \cite{xie2021segformer}. The third category is segmentation algorithms for sea fog detection, including Unet \cite{zhu_sea_2019} and SFCNet \cite{huang_correlation_2021}.

\begin{table*}[ht]
	\centering
	\caption{Comparison of Different Detection Models on SFDD}
	\label{table3}
	\resizebox{\textwidth}{39mm}{
	\begin{tabular}{llllllll}
		\hline
		& \textbf{CSI}   & \textbf{mIoU}  & \textbf{Acc}   & \textbf{Precision} & \textbf{Recall} & \textbf{F1}    & \textbf{Kappa} \\ \hline
		DeepLabV3+ \cite{chen_encoder-decoder_2018}        & 0.559          & 0.747          & 0.940          & 0.813              & 0.641           & 0.717          & 0.684          \\
		SegNet \cite{badrinarayanan_segnet_2017}            & 0.478          & 0.699          & 0.925          & 0.738              & 0.576           & 0.647          & 0.606          \\
		RefineNet \cite{lin_refinenet_2017}         & 0.500          & 0.713          & 0.931          & 0.778              & 0.584           & 0.667          & 0.629          \\
		PAN \cite{li_pyramid_nodate}              & 0.527          & 0.730          & 0.937          & {\ul 0.841}        & 0.585           & 0.690          & 0.657          \\
		BiSegNet \cite{yu_bisenet_2018}          & 0.563          & 0.749          & 0.941          & 0.831              & 0.635           & 0.720          & 0.688          \\
		DenseASPP \cite{yang_denseaspp_2018}        & 0.524          & 0.727          & 0.935          & 0.808              & 0.599           & 0.688          & 0.653          \\
		Unet \cite{zhu_sea_2019}             & 0.565          & 0.747          & 0.943          & 0.804              & 0.696           & 0.746          & 0.714          \\
		SFCNet \cite{huang_correlation_2021}           & 0.571          & 0.748          & 0.940          & 0.815              & 0.602           & 0.692          & 0.704          \\
		Swin-Transformer \cite{liu2021swin} & {\ul 0.597}    & {\ul 0.768}    & {\ul 0.943}    & 0.798              & {\ul 0.703}     & {\ul 0.748}    & {\ul 0.716}    \\
		Segmentor \cite{strudel2021segmenter}        & 0.545          & 0.740          & 0.940          & 0.833              & 0.607           & 0.702          & 0.673          \\
		Segformer \cite{xie2021segformer}        & 0.565          & 0.752          & 0.943          & \textbf{0.867}     & 0.619           & 0.722          & 0.692          \\
		\textbf{DB-SFNet} & \textbf{0.627} & \textbf{0.784} & \textbf{0.946} & 0.775              & \textbf{0.767}  & \textbf{0.771} & \textbf{0.740} \\ \hline
	\end{tabular}
}
\end{table*}

\subsubsection{Quantitative study}

\begin{figure*}[htbp]
	\centering
	\includegraphics[width=\linewidth]{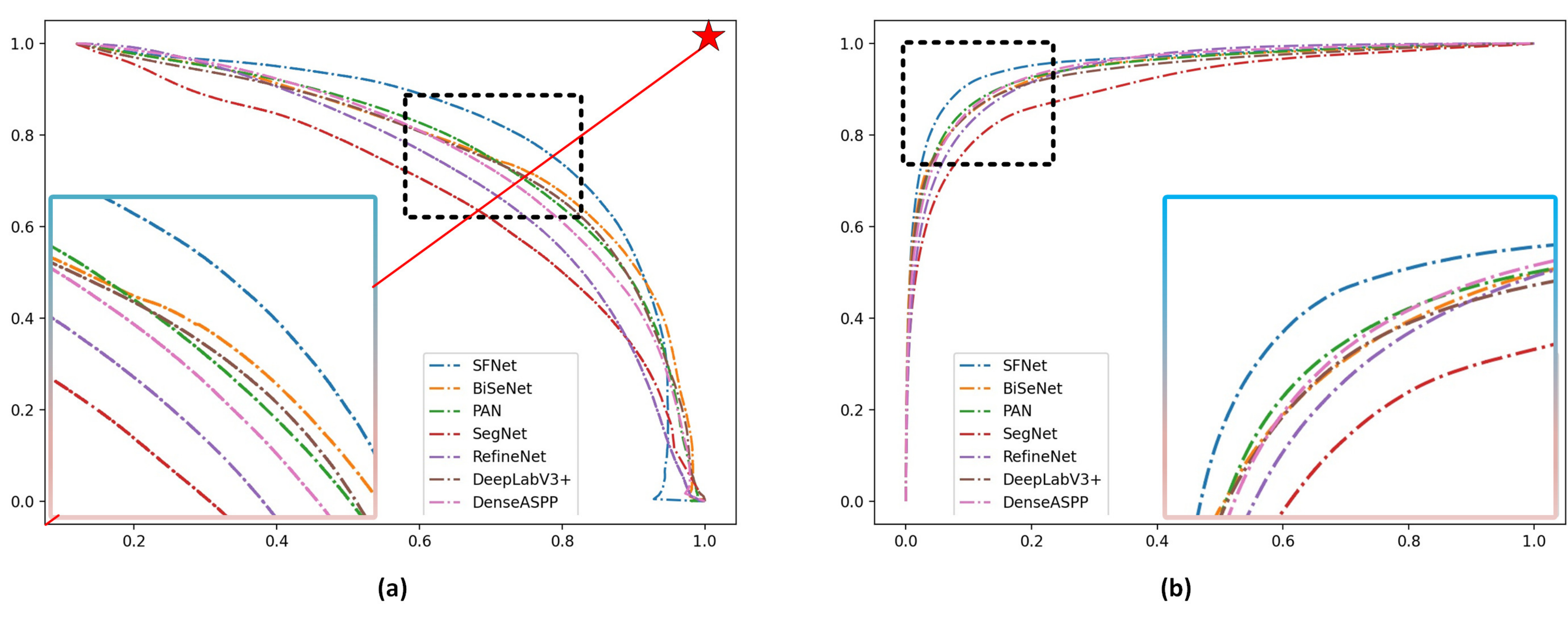}
	\caption{PR curves and ROC curves of different detection algorithms. (a) PR curves. PR curve plots the precision of a model as a function of its recall. The curve of DB-SFNet is closer to the upper left corner compared with that of other detection models. (b) ROC curves. The ROC curve is used to make a balance between the benefits, i.e., true positives, and costs, i.e., false positives. The area between the ROC curve of DB-SFNet and the horizontal axis is the largest.}
	\label{Fig5}
\end{figure*}

DB-SFNet and the other comparison models are tested on 208 images in the test set. Table \ref{table3} presents the quantitative results in CSI, mIoU, Accuracy, Precision, Recall, F1-score, and Kappa. It should be noted that there are a few images without sea fog or with low-coverage sea fog in the test set. Yang et al. \cite{yang_cdnet_2019} indicated that a low foreground coverage of less than 5\% may cause an apparent reduction in the evaluation metrics. Some scholars prefer removing these samples while evaluating the performance of a model. Nevertheless, images without sea fog or small sea fog coverage are common real scenarios. Therefore, we chose to retain these samples for evaluation.

The highest score in the segmentation method are bold and second-highest score are underlined in Table \ref{table2}. Among the deep learning-based detection methods, DB-SFNet shows better performance than the other algorithms in terms of all the parameters, except for precision. DB-SFNet is the only model with CSI above 0.6. The comprehensive evaluation metrics F1 score, Kappa, and mIoU of DB-SFNet are also significantly higher than those of the other models. The results of the quantitative analysis show that: 1. Deep learning models can distinguish sea fog from the other areas; 2. Statistical features are effective in sea fog detection; 3. The proposed DB-SFNet exhibits advantages in sea fog detection owing to using CNN to process statistical features extracted from prior knowledge.

The output of the deep learning-based sea fog detection models is essentially the probability of sea fog at each pixel. Pixels with a probability greater than the critical value are detected as fog areas. The evaluation indicators in Table \ref{table2} are calculated when the critical value is 0.5. To evaluate our models independently from the critical value, Fig. \ref{Fig5} shows the Precision-Recall curve (PR curve) and Receiver Operating Characteristic curve (ROC curve) of different semantic segmentation models.

PR curve plots the precision of a model as a function of its recall. A given critical value corresponds to a single point in PR space, and by varying the critical value, a PR curve can be obtained: while decreasing t from 1.0 to 0.0, an increasing number of instances is predicted as positive, causing the recall to increase, whereas precision may increase or decrease. Point (1,1) in the PR curve represents a classifier that obtains 100\% precision and sensitivity, which is the ideal PR curve point. Hence, the closer the PR curve is to the upper right corner, the better is the performance of sea fog detection\cite{tharwat_classification_2020}. As shown in Fig. \ref{Fig5}(a), the curve of DB-SFNet is closer to the upper left corner of the figure than the other detection models, confirming that DB-SFNet exhibits better detection performance.

ROC curve is a two-dimensional graph, with the TP plotted along the y-axis and FP plotted along the x-axis. The ROC curve is used to balance the benefits, i.e., true positives and costs, i.e., false positives. Similar to the PR curve, the ROC curve is generated by changing the critical value on the confidence score; hence, each critical value generates only one point in the ROC curve. The larger the area between the ROC curve and the horizontal axis, the better the performance of the corresponding detection algorithm \cite{tharwat_classification_2020}. Among the seven ROC curves shown in Fig. \ref{Fig5}(b), the area between the ROC curve of DB-SFNet and the horizontal axis is the largest, implying that DB-SFNet has better detection performance.

\subsubsection{Qualitative study}

\begin{figure*}[htbp]
	\centering
	\includegraphics[]{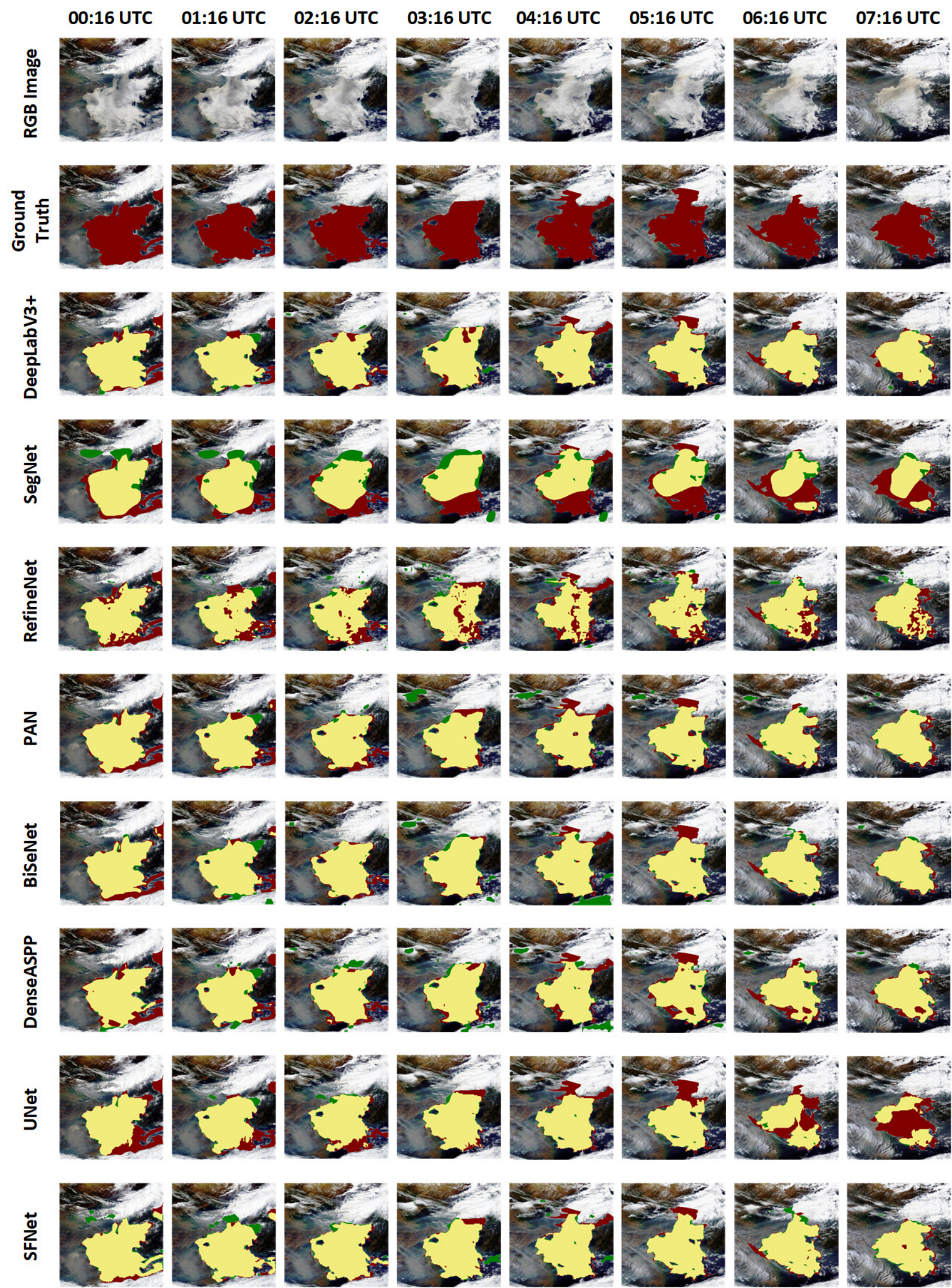}
	\caption{The detection results of different detection models on a complete sea fog event with mixed clouds and fog. The selected sea fog event occurred on 03 June 2011. The images record the changes in sea fog every hour from 00:16 UTC to 07:16 UTC. The first and second lines present the original images and ground truth. The rest are the sea fog detection results of different models. The correct pixels for foreground detection are indicated in yellow. False-negative and false-positive pixels are indicated by green and red, respectively.}
	\label{Fig6}
\end{figure*}

\begin{figure*}[htbp]
	\centering
	\includegraphics[width=0.90\textwidth]{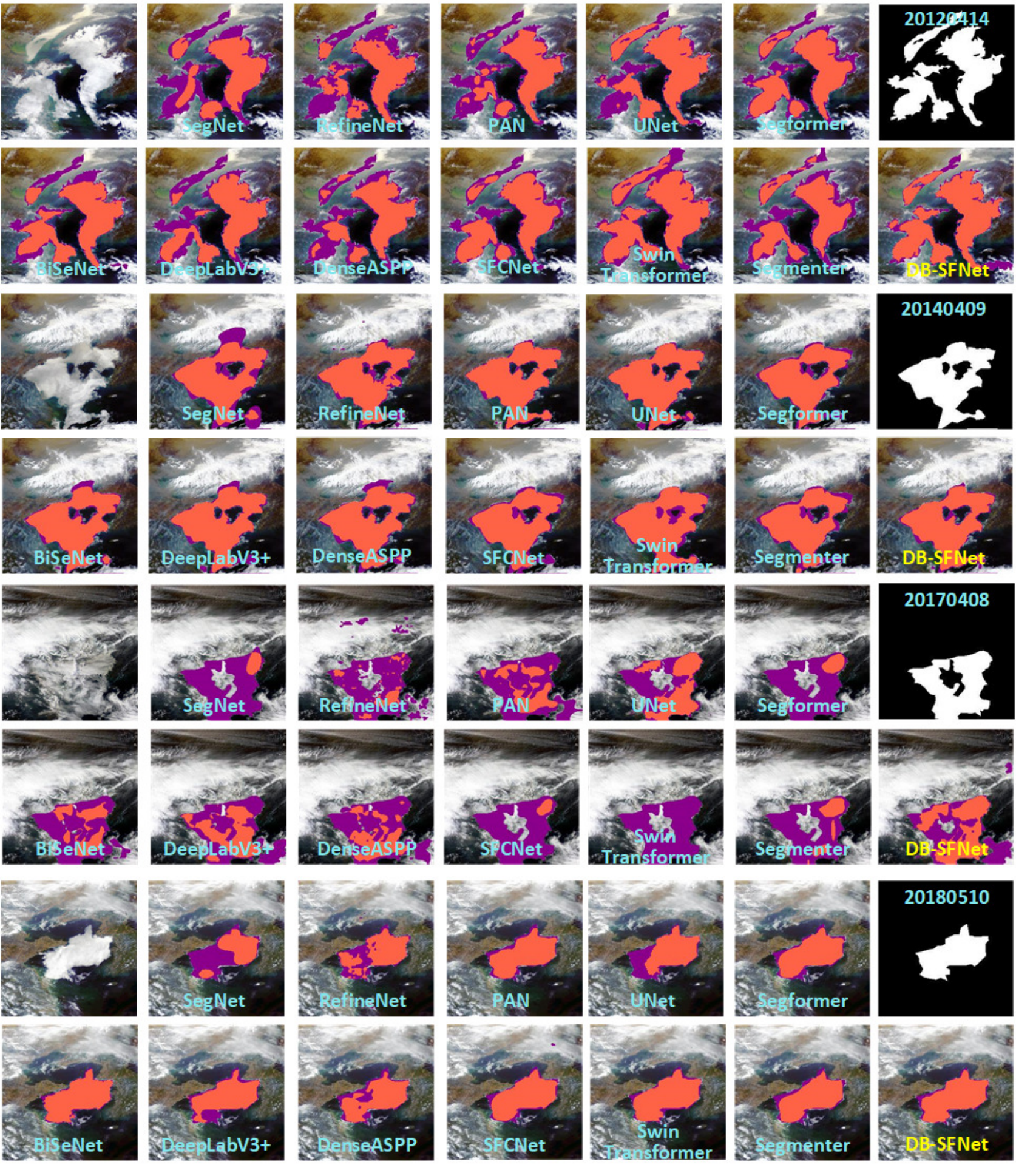}
	\caption{The detection results of different detection models on occurred sea fog events with mixed clouds and fog. The correct pixels for foreground detection in each subfigure are red, while the correct pixels for background detection are black. False-negative and false-positive pixels are indicated in purple.}
	\label{Fig7}
\end{figure*}

\begin{figure*}[htbp]
	\centering
	\includegraphics[]{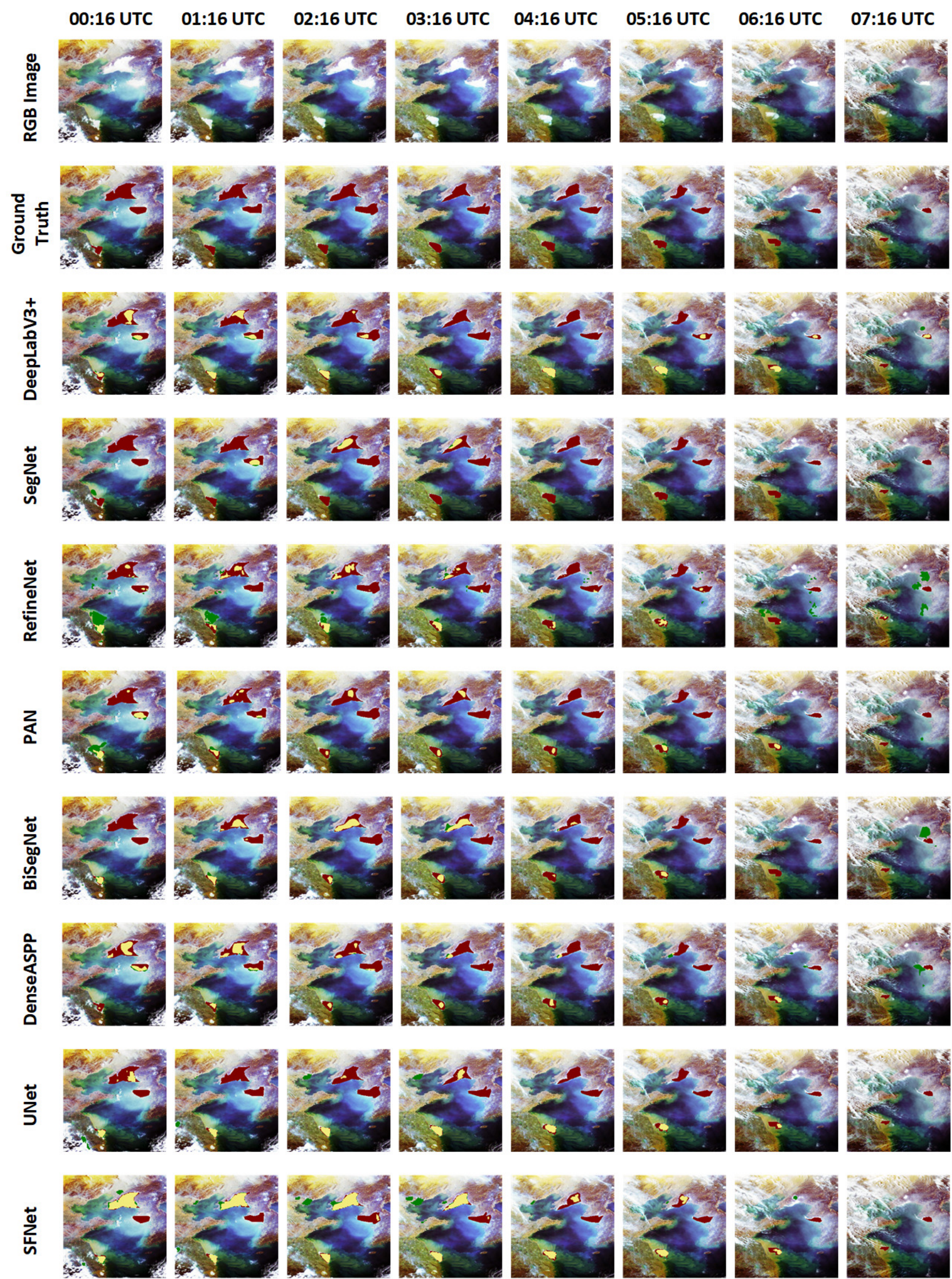}
	\caption{The detection results of different detection models on a complete sea fog event without clouds. The selected sea fog event occurred on 19 March 2019. The images record the changes in sea fog every hour from 00:16 UTC to 07:16 UTC. The first and second lines present the original images and ground truth. The rest are the sea fog detection results of different models. The correct pixels for foreground detection are indicated in yellow. False-negative and false-positive pixels are indicated by green and red, respectively.}
	\label{Fig8}
\end{figure*}

The DB-SFNet is not only superior to the existing models in terms of statistical indicators but is also closer to ground truth in randomly selected cases. We performed a visual analysis of some samples from the test set. Depending on the different foreground coverage, sea fog images of the test set can be divided into two categories: with and without cloud interference. For visualization, we randomly selected a complete sea fog event with mixed clouds and fog from 00:16 UTC to 07:16 UTC. The qualitative comparison chart is shown in Fig. \ref{Fig6}. In addition, we randomly selected some images with mixed clouds and no time continuity and visualized them in Fig \ref{Fig7}. It can be found from Fig. \ref{Fig6} and Fig. \ref{Fig7} that DB-SFNet can effectively improve the success rate of fog area detection. We also select a sea fog event without clouds and show the results in Fig. \ref{Fig8}. In this case, the results of models exhibit slight visual difference.

\subsection{Case Study on Different Seasons}
\begin{figure*}[htbp]
	\centering
	\includegraphics[width=0.9\textwidth]{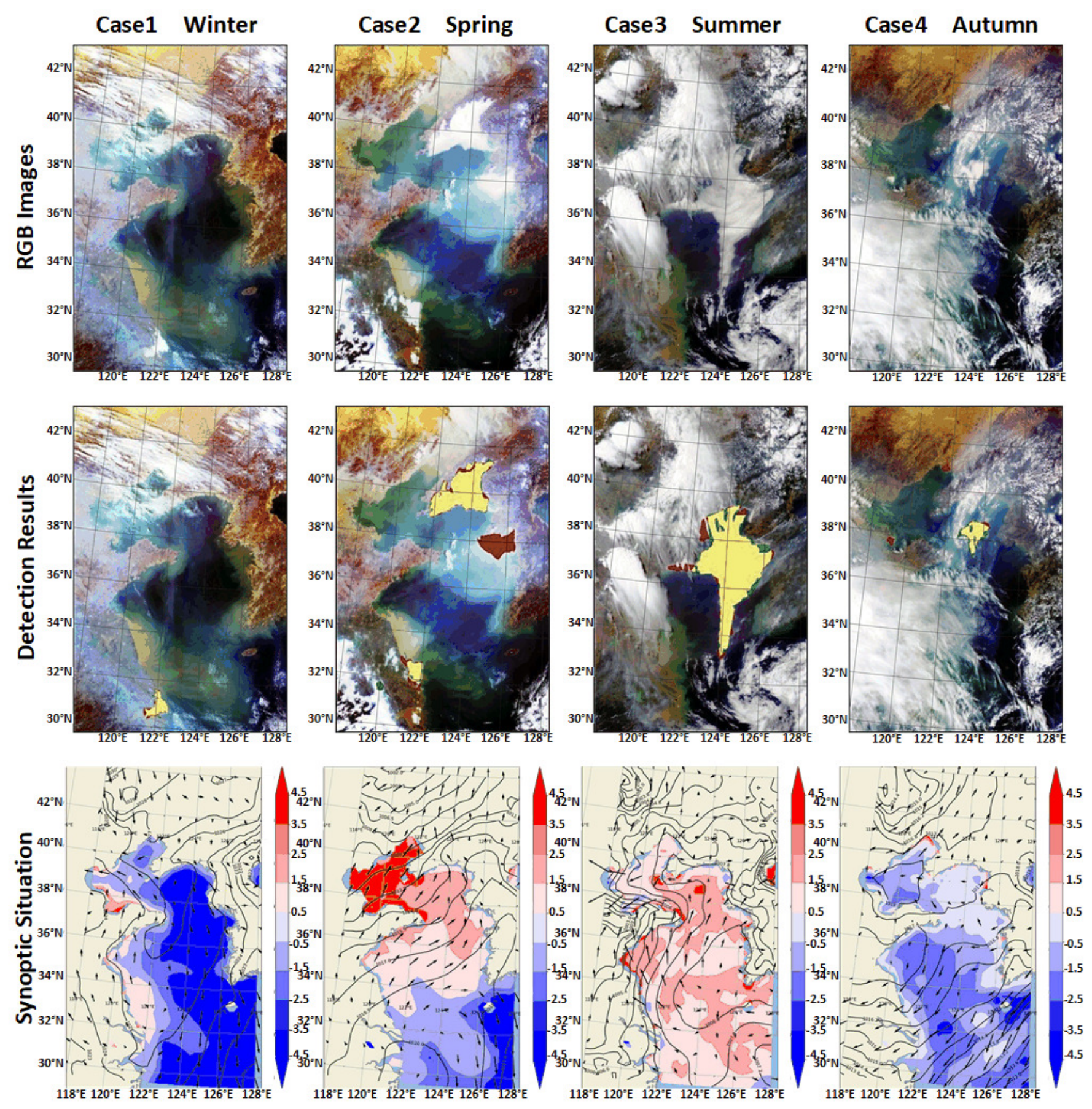}
	\caption{Four sea fog events across four seasons. Each column represents a sea fog event. RGB images are the true color images collected by GOCI. In the detection result, the yellow pixels are the correct areas, the red area corresponds to undetected sea fog that occurred, while the green areas correspond to erroneous reports. The synoptic situation shows the wind field, mean sea level pressure field, and the difference between SAT and SST. (a) Case 1: Winter sea fog event at 05:16 UTC on 24 January 2019. (b) Case 2: Spring sea fog event at 00:16 UTC on 19 March 2019. (c) Case 3: Summer sea fog event at 07:16 UTC on 04 June 2019. (d) Case 4: Autumn sea fog event at 05:16 UTC on 22 October 2019.}
	\label{Fig9}
\end{figure*}

The sea fog in the Yellow and Bohai Sea generally belongs to the type of advection fog that forms as warm/moist air passing over colder water. The formation of warm/moist air is closely related to seasonal variations. Thus, sea fog generation is also influenced by seasonal variations. To examine the influences of seasonal variations on sea fog generation, we applied DB-SFNet to detect sea fog across the four seasons of spring, summer, autumn, and winter in 2019. The four cases shown in Fig. \ref{Fig9} were selected to examine the sea fog detection across different seasons. We chose January, March, June, and October to represent winter, spring, summer, and autumn.

It can be seen from the synoptic situation in Fig. \ref{Fig9} that the sea temperature difference across four cases is considerably different. In Case 2 and Case 3 (spring and summer), the air temperature is higher than the sea surface temperature. Thus sea fog is more likely to occur. As shown in Fig. \ref{Fig9} (b) and (c), sea fog areas are prominent in these two cases. In contrast, the sea surface temperature is higher in Case 1 and Case 4, which is not conducive to sea fog formation. The fog area in autumn and winter is often smaller than in spring and summer. 

DB-SFNet generates the detection results in Fig. \ref{Fig9}. The yellow pixels are the correctly detected areas (TP), the red areas correspond to undetected sea fog that occurred (FP), and the green areas are erroneous reports (FN). It can be seen from the detection results that the detection ability of DB-SFNet is not affected by seasonal variations. Across different seasons and weather conditions, the sea fog area detected by DB-SFNet is consistent with the actual sea fog area. Smaller sea fog areas appear in autumn and winter (Case 1 and Case 4), and larger sea fog areas appear in spring and summer (Case 2 and Case 3) are successfully detected.

\subsection{Ablation Experiments}

This paper proposes a novel dataset (SFDD) and model (DB-SFNet) to improve the performance of sea fog detection. To validate the proposed dataset and model, we set up two ablation experiments Our ablation studies aim to answer the following questions:

\subsubsection{Can training on a long period of data improve the model's generalization ability?}

The SFDD data covers ten years, and the sample types are abundant. To prove that the model trained on a long time span of data is less prone to overfitting, we trained DeeplabV3+ and DB-SFNet on data with different periods (the longer the period, the closer the data distribution is to the real distribution). The following variants of SFDD are implemented to verify that SFDD can reduce the risk of overfitting. The test set is data from 2019-2020.
\begin{itemize}[]
	\item Model-SFDD-2: Data from 2011 to 2012 is used for training;
	
	\item Model-SFDD-4: Data from 2011 to 2014 is used for training;
	
	\item Model-SFDD-6: Data from 2011 to 2016 is used for training;
	
	\item Model-SFDD-8: Data from 2011 to 2018 is used for training.
\end{itemize}

\begin{table}[]
	\centering
	\caption{Ablation result on time span}
	\label{table4}
	\begin{tabular}{llll}
		\hline
		\textbf{}                                                             & \textbf{mIoU} & \textbf{F1-Score} & \textbf{Kappa} \\ \hline
		\multicolumn{1}{l|}{{\color[HTML]{231F20} \textbf{DeeplabV3+-SFDD2}}} & 0.6607        & 0.5167            & 0.5427         \\
		\multicolumn{1}{l|}{{\color[HTML]{231F20} \textbf{DeeplabV3+-SFDD4}}} & 0.6939        & 0.5801            & 0.6017         \\
		\multicolumn{1}{l|}{{\color[HTML]{231F20} \textbf{DeeplabV3+-SFDD6}}} & 0.6984        & 0.5892            & 0.6113         \\
		\multicolumn{1}{l|}{{\color[HTML]{231F20} \textbf{DeeplabV3+-SFDD8}}} & 0.7187        & 0.6261            & 0.6464         \\ \hline
		\multicolumn{1}{l|}{{\color[HTML]{231F20} \textbf{DB-SFNet-SFDD2}}}  & 0.7033        & 0.5986            & 0.6206         \\
		\multicolumn{1}{l|}{{\color[HTML]{231F20} \textbf{DB-SFNet-SFDD4}}}  & 0.7163        & 0.6211            & 0.6408         \\
		\multicolumn{1}{l|}{\textbf{DB-SFNet-SFDD6}}                         & 0.7261        & 0.6393            & 0.6593         \\
		\multicolumn{1}{l|}{\textbf{DB-SFNet-SFDD8}}                         & 0.7597        & 0.6944            & 0.7101         \\ \hline
	\end{tabular}
\end{table}

\begin{table}[]
	\centering
	\caption{Ablation result on the scale of feature fusion}
	\label{table5}
	\begin{tabular}{llll}
		\hline
		\textbf{}                                                                 & \textbf{mIoU} & \textbf{F1Score} & \textbf{Kappa} \\ \hline
		\multicolumn{1}{l|}{{\color[HTML]{231F20} \textbf{DB-SFNet (Base)}}}     & 0.7466        & 0.7209           & 0.7115         \\
		\multicolumn{1}{l|}{{\color[HTML]{231F20} \textbf{DB-SFNet (block1)}}}   & 0.7707        & 0.7512           & 0.7205         \\
		\multicolumn{1}{l|}{{\color[HTML]{231F20} \textbf{DB-SFNet (block2)}}}   & 0.7743        & 0.7567           & 0.7259         \\
		\multicolumn{1}{l|}{{\color[HTML]{231F20} \textbf{DB-SFNet (block3)}}}   & 0.7747        & 0.7564           & 0.7262         \\
		\multicolumn{1}{l|}{{\color[HTML]{231F20} \textbf{DB-SFNet (block4)}}}   & 0.7765        & 0.7595           & 0.7291         \\
		\multicolumn{1}{l|}{{\color[HTML]{231F20} \textbf{DB-SFNet (complete)}}} & 0.7835        & 0.7707           & 0.7399         \\ \hline
	\end{tabular}
\end{table}

Training on SFDD can improve the model's generalization ability. As shown in Table \ref{table4}, the longer the period of the training set, the higher the detection scores of DeepLabV3+ and DB-SFNet, partly because the training set's data distribution is closer to the actual observation data distribution when the period is longer. In addition, by comparing the detection scores of DB-SFNet and DeepLabV3+, it can be found that the detection performance of DB-SFNet is always better than DeepLabV3+ when the data set is fixed, confirming the advantages of DB-SFNet.

\subsubsection{Are the discriminative features extracted from the statistical domain useful? If it is useful, at what scale does the fusion of statistical features perform better?} 

Unlike the current advanced sea fog detection methods, DB-SFNet combines the discriminative features from the statistical domain. To explore whether to introduce statistical features and at which scale is better to fuse, we designed the following variants of DB-SFNet.

\begin{itemize}[]
	\item DB-SFNet (Base): Basic model without statistical features.
	\item DB-SFNet (block1): Fuse statistical features after the first up sampling layer.
	\item DB-SFNet (block2): Fuse statistical features after the second up sampling layer.
	\item DB-SFNet (block3): Fuse statistical features after the third up sampling layer.
	\item DB-SFNet (block4): Fuse statistical features after the fourth up sampling layer.
	\item DB-SFNet (overall): Overall DB-SFNet model.
\end{itemize}

As shown in Table \ref{table5}, models that fuse statistical features at each scale have the best performance in sea fog detection. When the model does not add statistical features (DB-SFNet(Base)), the effects of the three sea fog detection indicators are not ideal. Combining statistical features at any scale can effectively improve the detection index. It is demonstrated that extracting features from the statistical domain is valuable. 

\subsubsection{How much improvement does each statistical feature contribute?}

On the basic architecture of DB-SFNet, we design ablation studies that only introduce one statistical feature. These studies demonstrate the contribution of each statistical feature. As shown in Table \ref{table6}, the introduction of each statistical feature has a positive effect on sea fog detection.

\begin{table}[ht]
	\centering
	\caption{Ablation experiments on statistical features}
	\label{table6}
	\begin{tabular}{llll}
		\hline
		& \textbf{mIou}   & \textbf{F1-Score} & \textbf{Kappa}  \\ \hline
		Base                & 0.7346          & 0.6975           & 0.6641          \\
		Base+$F^{mean}$          & 0.7381          & 0.7027           & 0.6696          \\
		Base+$F^{variance}$      & 0.7453          & 0.7118           & 0.6808          \\
		Base+$F^{homogeneity}$   & 0.7420          & 0.7075           & 0.6757          \\
		Base+$F^{contrast}$      & 0.7406          & 0.7057           & 0.6736          \\
		Base+$F^{entropy}$       & 0.7411          & 0.7064           & 0.6740          \\
		Base+$F^{dissimilarity}$ & 0.7496          & 0.7183           & 0.6876          \\
		Base+$F^{energy}$        & 0.7508          & 0.7203           & 0.6896          \\
		Base+$F^{correlation}$   & 0.7352          & 0.6994           & 0.6652          \\
		\textbf{DF-SFNet}   & \textbf{0.7835} & \textbf{0.7707}  & \textbf{0.7399} \\ \hline
	\end{tabular}
\end{table}

\subsubsection{How much improvement does the feature selection module contribute?}

To prove that the feature selection module ($FS$) has a positive effect, we separately trained the DB-SFNet with feature selection and that without feature selection. As shown in Table\ref{table7}, the performance of DB-SFNet without feature selection is already higher than most of the comparison models. And DB-SFNet using feature selection can achieve better performance. We believe that the feature selection block reduces the decoding complexity of visual and statistical features, thereby significantly improving model performance.

\begin{table}[ht]
	\centering
	\caption{Ablation experiments on feature selection}
	\label{table7}
	\begin{tabular}{llll}
		\hline
		& \textbf{mIou}   & \textbf{F1Score} & \textbf{Kappa}  \\ \hline
		DF-SFNet without FS & 0.757           & 0.7293           & 0.7113          \\
		\textbf{DF-SFNet}   & \textbf{0.7835} & \textbf{0.7707}  & \textbf{0.7399} \\ \hline
	\end{tabular}
\end{table}

\subsection{Backwhirling Application in Real World}

Sea fog in the Yellow Sea and the Bohai Sea mostly appears on China's east coast and the west of the Korean Peninsula due to the temperature difference between the sea and land. Therefore, we selected two sea fog incidents that caused significant losses from the sea fog reports recorded in China and South Korea as the analysis cases. To understand the causes behind these two sea fog events, we additionally used the fifth-generation global climate reanalysis data from ECMWF (ERA5) to analyze the climatic conditions in the areas where sea fog occurred. The selected meteorological elements include 10m u-component of wind, 10m v-component of wind, sea surface temperature (SST), surface air temperature (SAT), and mean sea level pressure.

\begin{figure}[ht]
	\centering
	\includegraphics[width=0.45\textwidth]{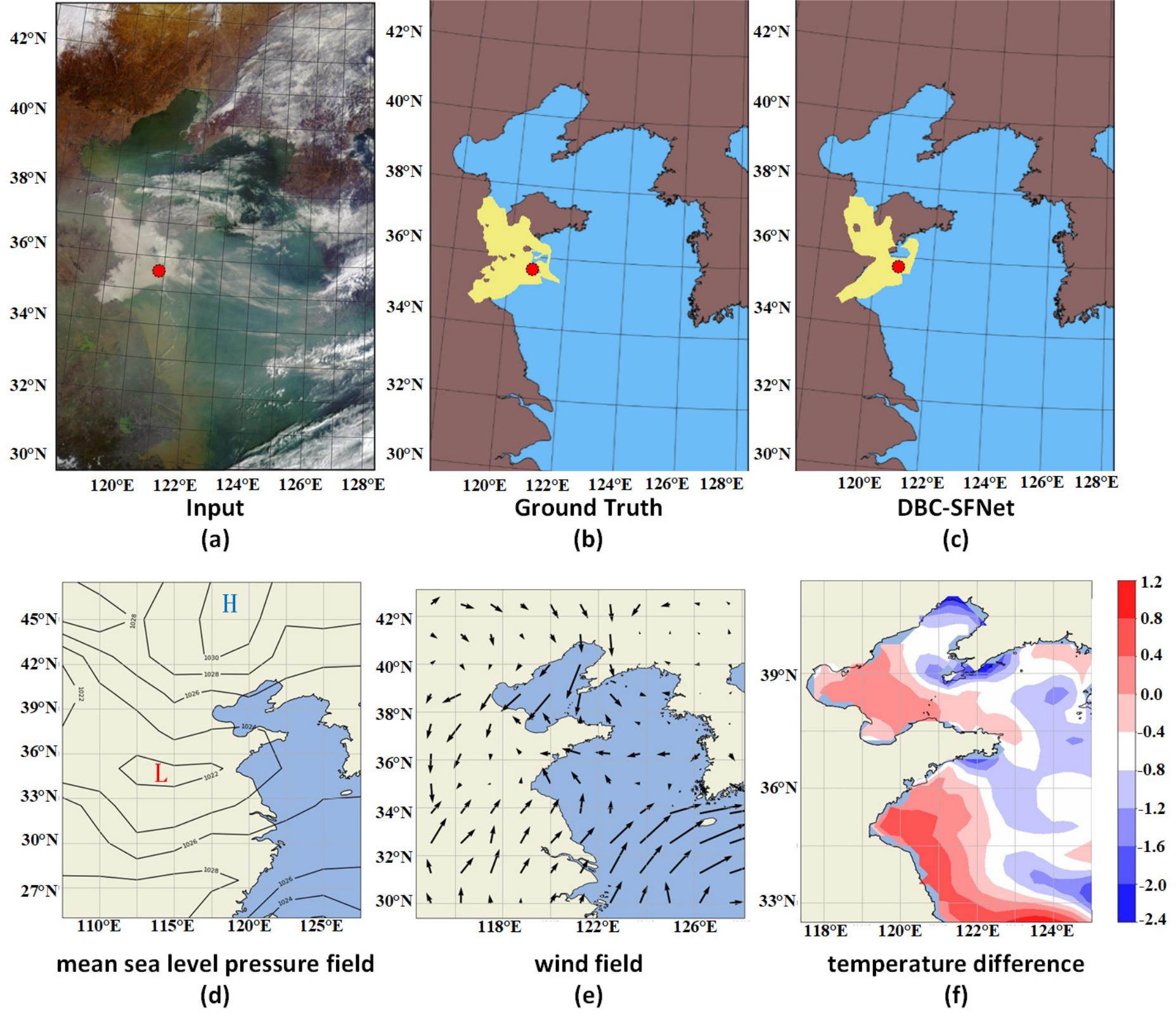}
	\caption{Case 1: A sea fog event on 27 February 2016. This event caused a ship collision at approximately 18 n miles south of Chaolian Island, Qingdao (Red point in (a), (b) and (c)). (a) GOCI true color image. The acquisition time is 00:16 UTC on 27 February 2016. (b) Actual sea fog area. The sea fog is indicated in yellow. (c) Sea fog area detected by DB-SFNet. The sea fog is indicated in yellow. (d) Mean sea level pressure field provided by ERA5. Qingdao is in the east of the low-pressure system. (e) Wind field provided by ERA5. Southerly wind prevails in the Shandong Peninsula (f) Temperature difference (SAT-SST). The temperature difference between the ocean and the air at the accident site is greater than 0.5$^{\circ}$C.}
	\label{Fig10}
\end{figure}

Case 1: Fig. \ref{Fig10} shows a sea fog event at 00:16 UTC on 27 February 2016. In this case, sea fog appears in the Yellow Sea on the south side of the Shandong Peninsula (yellow area in Fig. \ref{Fig10}(b)). In this case, a bulk carrier and a fishing boat collided (redpoint in Figure \ref{Fig10}(a)). Eight people died, two were missing, and the direct economic loss was approximate 1.2 million CNY. On 27 February, a low-pressure system was formed in inland China and the western Yellow Sea (shown in Fig. \ref{Fig10}(d)). The existence of the low-pressure system makes the Yellow Sea affected by the southeasterlies wind (shown in Figure \ref{Fig10}(e)). The warm and humid southerly air is transported to the colder sea, causing the temperature difference between the sea surface and surface air in the accident sea area above 0.5$^{\circ}$C (shown in Fig. \ref{Fig10}(f)). The temperature difference and humidity meet the conditions for sea fog. Fig. \ref{Fig10}(c) shows the sea fog detection result from DB-SFNet. The detection model proposed in this paper successfully detects the sea fog event. Specifically, the fog area detected by DB-SFNet is almost the same as the actual fog area, and only the edge of the fog area has individual false negatives. The detection results appropriately correspond to the accident point.

\begin{figure}[htbp]
	\centering
	\includegraphics[width=0.45\textwidth]{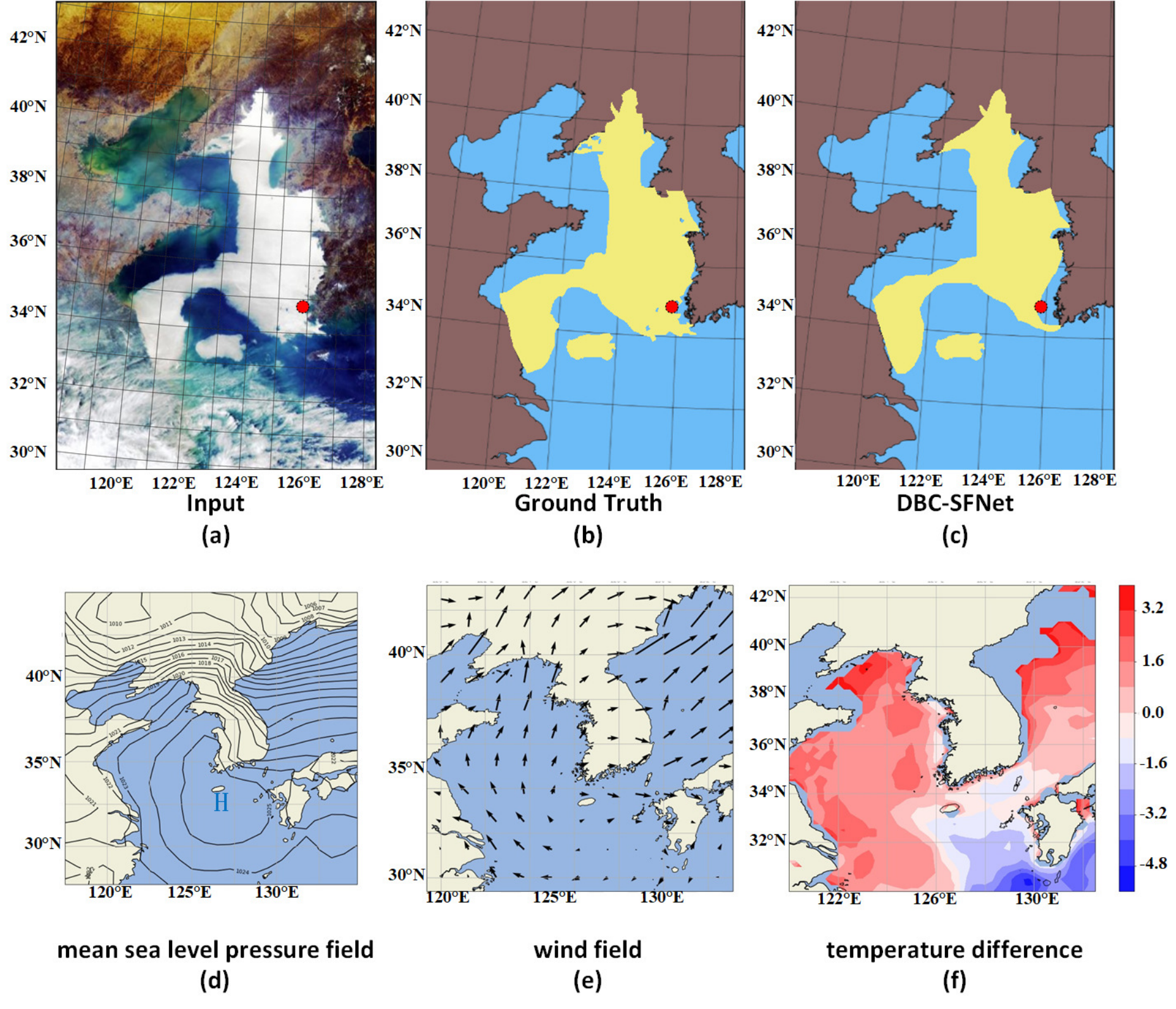}
	\caption{Case 2: A sea fog event on 25 March 2018. This event caused a Korean ship to strike a massive rock in the southwest islands of South Korea (Redpoint in (a), (b), and (c)). (a) GOCI true-color image. The acquisition time is 06:16 UTC on 25 March 2018. (b) Actual sea fog area. The sea fog is indicated in yellow. (c) Sea fog area detected by DB-SFNet. The sea fog is indicated in yellow. (d) Mean sea level pressure field provided by ERA5. A high pressure system is formed on the south side of the Korean peninsula. (e) Wind field provided by ERA5. Southerly wind prevails in the Korean Peninsula (f) Temperature difference (SAT-SST). The temperature difference between the ocean and the air at the accident site is greater than 0.5$^{\circ}$C.}
	\label{Fig11}
\end{figure}

Case 2: Fig. \ref{Fig11} shows a sea fog event at 06:16 UTC on 25 March 2016. The sea fog event occurred at the Yellow Sea, and West Korea Bay (indicated by the yellow area in Fig.\ref{Fig11}(b)). As the sea fog reduced visibility, a passenger ship carrying about 160 people was grounded on rocks in waters off the coast of Sinan in South Jeolla Province (indicated by the red point in Fig. \ref{Fig11}(a)). On 25 March, a high-pressure system was formed on the south of the Korean peninsula (as shown in Fig. \ref{Fig11}(d)). The sea fog area was at the rear of the offshore high-pressure region, causing southerly wind in the Yellow Sea (as shown in Fig. \ref{Fig11}(e)). The warm and humid southerly air is transported to the colder sea, thus causing the temperature between the sea surface and surface air in the accident sea area to be greater than 0.5$^{\circ}$C (as shown in Fig. \ref{Fig11}(f)). The temperature difference and humidity meet the conditions for sea fog formation. Fig. \ref{Fig11}(c) depicts the sea fog detection result using DB-SFNet. DB-SFNet effectively detects the sea fog event. Specifically, the fog region detected by DB-SFNet is the actual fog area, and only the periphery of the fog area contains individual false negatives. The detection results appropriately correspond to the accident spot.

\subsection{Complexity Analysis}

\begin{figure*}[htbp]
	\centering
	\includegraphics[width=\textwidth]{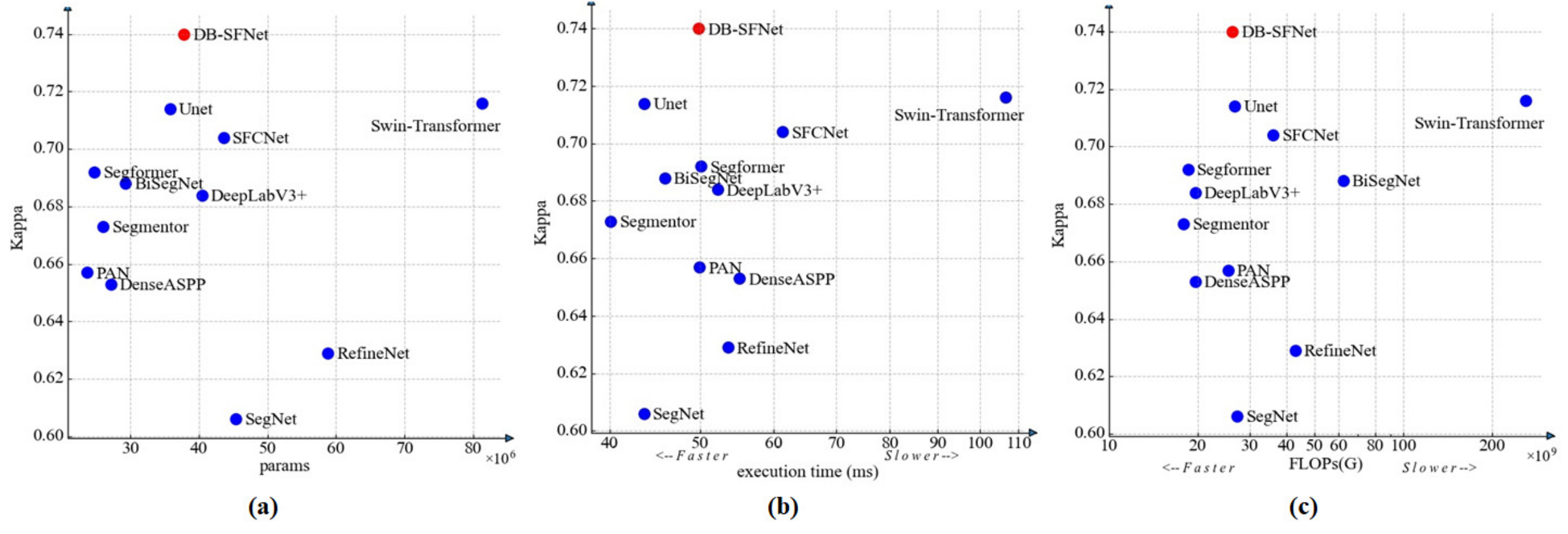}
	\caption{Complexity Analysis. (a) Comparison of $Kappa$ and model parameters; (b) Comparison of $Kappa$ and inference time; (c) Comparison of $Kappa$ and FLOPs.}
	\label{Fig12}
\end{figure*}

In this part, we analyze the complexity of the proposed algorithm according to parameter amount, computational complexity and execution time. The results are illustrated in Fig. \ref{Fig12}.

As shown in Fig. \ref{Fig12} (a), our DB-SFNet achieves high performance with an affordable parameter number, compared with other methods. With regard of the execution time in Fig. \ref{Fig12} (b), the proposed method takes only 0.058 seconds to process an image, which is superior to a lot of comparative state-of-the-art methods. Comparisons about FLOPs and performance are shown in Fig. \ref{Fig12} (c), which demonstrate that the proposed DF-SFNet has a better tradeoff between FLOPs and performance. 

\begin{table*}[ht]
	\centering
	\caption{Comparison of Different Detection Models on Himawari-8 satellite}
	\label{table8}
	\resizebox{\textwidth}{39mm}{
		\begin{tabular}{llllllll}
			\hline
			& \textbf{CSI}    & \textbf{mIoU}   & \textbf{Acc}    & \textbf{Precision} & \textbf{Recall} & \textbf{F1}    & \textbf{Kappa}  \\ \hline
			DeepLabV3+ \cite{chen_encoder-decoder_2018}       & 0.761           & 0.8717          & 0.9834          & \textbf{0.9456}    & 0.7958          & 0.8643         & 0.8555          \\
			SegNet \cite{badrinarayanan_segnet_2017}           & 0.7502          & 0.8658          & 0.9824          & 0.9246             & 0.7991          & 0.8573         & 0.8479          \\
			RefineNet \cite{lin_refinenet_2017}        & 0.6638          & 0.8179          & 0.9734          & 0.8023             & 0.7936          & 0.7979         & 0.7837          \\
			PAN \cite{li_pyramid_nodate}              & 0.749           & 0.8653          & 0.9825          & 0.9391             & 0.7872          & 0.8565         & 0.8472          \\
			BiSeNet \cite{yu_bisenet_2018}          & 0.7564          & 0.8692          & 0.9829          & 0.9336             & 0.7994          & 0.8613         & 0.8523          \\
			DenseASPP \cite{yang_denseaspp_2018}        & 0.7449          & 0.8625          & 0.9811          & 0.8798             & 0.8293          & 0.8538         & 0.8437          \\
			Unet \cite{zhu_sea_2019}             & 0.7567          & 0.8694          & 0.9831          & {\ul 0.9443}       & 0.792           & 0.8615         & 0.8526          \\
			SFCNet \cite{huang_correlation_2021}           & 0.7488          & 0.8643          & 0.981           & 0.8586             & \textbf{0.8541} & 0.8563         & 0.8462          \\
			Swin Transformer \cite{liu2021swin}  & 0.7755          & 0.8793          & 0.9841          & 0.9246             & 0.8278          & 0.8735         & 0.8651          \\
			Segmenter \cite{strudel2021segmenter}        & 0.6639          & 0.8187          & 0.9748          & 0.8504             & 0.7517          & 0.798          & 0.7846          \\
			SegFormer \cite{xie2021segformer}        & {\ul 0.779}     & {\ul 0.8813}    & {\ul 0.9845}    & 0.934              & 0.8244          & {\ul 0.8758}   & {\ul 0.8676}    \\
			\textbf{DB-SFNet} & \textbf{0.7953} & \textbf{0.8901} & \textbf{0.9857} & 0.9368             & {\ul 0.8404}    & \textbf{0.886} & \textbf{0.8784} \\ \hline
		\end{tabular}
	}
\end{table*}

\subsection{Generalization Ability on Other Satellite Data}

To evaluate the generalization ability and effect of the proposed method, we have selected 77 sea fog images recorded by the Himawari-8 satellite. Among them, 56 images are used to train the models, and the remaining 21 images are used to test the performance. As shown in Table \ref{table8}, even though the model was trained on images from another satellite, our method compares favorably with most other methods in terms of all the conventional metrics. This shows that DB-SFNet has a satisfactory generalization ability and effect.

\section{Conculsion}

This paper proposes a sea fog detection dataset (SFDD) and a dual branch sea fog detection model (DB-SFNet) for robust, rapid, and accurate daytime sea fog detection. To the best of our knowledge, SFDD is the first sea fog detection dataset with an extended period, large sample number, and accurate labeling. SFDD is comprised of 1033 sea fog images collected by GOCI, containing all the observed sea fog events occured in the Yellow Sea and the Bohai Sea from 2010 to 2020. Moreover, we propose DB-SFNet that realize the rapid detection of sea fog from remote sensing. DB-SFNet contains a statistical feature extraction module that converts the visual representation to statistical domains and an encoder-decoder module for extracting discriminative features in the visual and statistical domains. By extracting the discriminative information in the statistical domain, the proposed model can improve the performance of sea fog detection, significantly reducing the false alarms of the cloud and fog mixing area. 

The main conclusions of this paper are as follows: (1) Large-scale SFDD can reflect real scenarios better. The seasonal variation and distribution of the SFDD is consistent with historical statistics, and it significantly reduces the risk of overfitting and improves the performance of fog detection models; (2) DB-SFNet outperforms advanced semantic segmentation algorithms in sea fog detection, especially, it can effectively detect sea fog in a mixed image of cloud and fog; (3) DB-SFNet that fuse statistical features at each scale have the best performance. It indicates that the statistical features facilitate the sea fog detection.

\bibliographystyle{IEEEtran}
\bibliography{mybibfile}

\end{document}